\documentclass[a4paper, 12 pt, twocolumn, final, conference]{ieeetran}
\IEEEoverridecommandlockouts
\usepackage{graphicx}
\usepackage{subcaption}
\usepackage{times}   % assumes new font selection scheme installed
\usepackage{amsmath} % assumes amsmath package installed
\usepackage{amssymb} % assumes amsmath package installed
\usepackage{avldefs} 
\usepackage{algorithm}
\usepackage{amsfonts}
\usepackage{algorithm, algorithmic}
\usepackage{array}
\usepackage{multirow}
\usepackage{microtype}
\usepackage{graphics}
\usepackage{booktabs} % for professional tables
\usepackage{url}
\usepackage{bm}  % for bold math

\usepackage[square, numbers]{natbib}

\usepackage{color}

\usepackage{booktabs}
\usepackage[normalem]{ulem}
\useunder{\uline}{\ul}{}

\title{\LARGE \bf
Enhancing Time Series Momentum Strategies \\ Using Deep Neural Networks
}

\author{Bryan Lim, Stefan Zohren, Stephen Roberts
\thanks{B. Lim, S. Zohren and S. Roberts are with the Department of Engineering Science and the Oxford-Man Institute of Quantitative Finance, University of Oxford, Oxford, United Kingdom (email: bryan.lim@eng.ox.ac.uk, zohren@robots.ox.ac.uk, sjrob@robots.ox.ac.uk).}
}

\makeatletter
\renewcommand*{\thetable}{\arabic{table}}
\renewcommand*{\thefigure}{\arabic{figure}}
\let\c@table\c@figure
\makeatother 

\usepackage{caption}
\begin{document}
\maketitle
\thispagestyle{plain}
\pagestyle{plain}
%
%%%%%%%%%%%%%%%%%%%%%%%%%%%%%%%%%%%%%%%%%%%%%%%%%%%%%%%%%%%%%%%%%%%%%%%%%%%%%%%%

\begin{abstract}
While time series momentum \cite{TimeSeriesMomentum} is a well-studied phenomenon in finance, common strategies require the explicit definition of both a trend estimator and a position sizing rule. In this paper, we introduce Deep Momentum Networks -- a hybrid approach which injects deep learning based trading rules into the volatility scaling framework of time series momentum.  The model also simultaneously learns both trend estimation and position sizing in a data-driven manner, with networks directly trained by optimising the Sharpe ratio of the signal. Backtesting on a portfolio of 88 continuous futures contracts, we demonstrate that the Sharpe-optimised LSTM improved traditional methods by more than two times in the absence of transactions costs, and continue outperforming when considering transaction costs up to 2-3 basis points. To account for more illiquid assets, we also propose a turnover regularisation term which trains the network to factor in costs at run-time.
\end{abstract}
%%%%%%%%%%%%%%%%%%%%%%%%%%%%%%%%%%%%%%%%%%%%%%%%%%%%%%%%%%%%%%%%%%%%%%%%%%%%%%%%

\section{Introduction}

Momentum as a risk premium in finance has been extensively documented in the academic literature, with evidence of persistent abnormal returns demonstrated across a range of asset classes, prediction horizons and time periods \cite{CenturyOfTrendFollowing, TwoCenturiesOfTrendFollowing, AHLMomentum}. Based on the philosophy that strong price trends have a tendency to persist, time series momentum strategies are typically designed to increase position sizes with large directional moves and reduce positions at other times. Although the intuition underpinning the strategy is clear, specific implementation details can vary widely between signals – with a plethora of methods available to estimate the magnitude of price trends \cite{WhichTrendIsYourFriend, TrendFilteringLyxor, AHLMomentum} and map them to actual traded positions \cite{TSMomAndVolScaling, DemystifyingTimeSeriesMomentum, VolTargeting}.

In recent times, deep neural networks have been increasingly used for time series prediction, outperforming traditional benchmarks in applications such as demand forecasting \cite{UberExtremeEvents}, medicine \cite{DiseaseAtlas} and finance \cite{DeepLOB}. With the development of modern architectures such as convolutional neural networks (CNNs) and recurrent neural networks (RNNs) \cite{DeepLearningBook}, deep learning models have been favoured for their ability to build representations of a given dataset \cite{RepresentationLearning} -- capturing temporal dynamics and cross-sectional relationships in a purely data-driven manner. The adoption of deep neural networks has also been facilitated by powerful open-source frameworks such as \texttt{TensorFlow} \cite{tensorflow} and \texttt{PyTorch} \citep{pytorch} -- which use automatic differentiation to compute gradients for backpropagation without having to explicitly derive them in advance. In turn, this flexibility has allowed deep neural networks to go beyond standard classification and regression models. For instance, the creation of hybrid methods that combine traditional time-series models with neural network components have been observed to outperform pure methods in either category \cite{M4Competition} – e.g. the exponential smoothing RNN \cite{ESRNN}, autoregressive CNNs \cite{AutoregressiveCNNs} and Kalman filter variants \cite{DeepStateSpaceModels, KVAE} -- while also making outputs easier to interpret by practitioners. Furthermore, these frameworks have also enabled the development of new loss functions for training neural networks, such as adversarial loss functions in generative adversarial networks (GANs) \cite{GANs}.

While numerous papers have investigated the use of machine learning for financial time series prediction, they typically focus on casting the underlying prediction problem as a standard regression or classification task \cite{EmpiricalAssetPricingViaML, XSectionalDeepMomentum, RamaContUniversality, DeepLOB, ConveNetEnsembles, AutoregressiveCNNs, LSTMFinancialTimeSeries} -- with regression models forecasting expected returns, and classification models predicting the direction of future price movements. This approach, however, could lead to suboptimal performance in the context time-series momentum for several reasons. Firstly, sizing positions based on expected returns alone does not take risk characteristics into account – such as the volatility or skew of the predictive returns distribution –- which could inadvertently expose signals to large downside moves. This is particularly relevant as raw momentum strategies without adequate risk adjustments, such as volatility scaling \cite{TSMomAndVolScaling}, are susceptible to large crashes during periods of market panic \cite{MomentumMoments, MomentumCrashes}. Furthermore, even with volatility scaling – which leads to positively skewed returns distributions and long-option-like behaviour \cite{SkewsMe, OptionProfileMomentum} -- trend following strategies can place more losing trades than winning ones and still be profitable on the whole -- as they size up only into large but infrequent directional moves. As such, \cite{TrendFollowersLoseMoreThanTheyGain} argue that the fraction of winning trades is a meaningless metric of performance, given that it cannot be evaluated independently from the trading style of the strategy. Similarly, high classification accuracies may not necessarily translate into positive strategy performance, as profitability also depends on the magnitude of returns in each class. This is also echoed in betting strategies such as the Kelly criterion \cite{KellyCriterion}, which requires both win/loss probabilities and betting odds for optimal sizing in binomial games. In light of the deficiencies of standard supervised learning techniques, new loss functions and training methods would need to be explored for position sizing -- accounting for trade-offs between risk and reward.

In this paper, we introduce a novel class of hybrid models that combines deep learning-based trading signals with the volatility scaling framework used in time series momentum strategies \cite{DemystifyingTimeSeriesMomentum, TimeSeriesMomentum} -- which we refer to as the \emph{Deep Momentum Networks} (DMNs). This improves existing methods from several angles. Firstly, by using deep neural networks to directly generate trading signals, we remove the need to manually specify both the trend estimator and position sizing methodology -- allowing them to be learnt directly using modern time series prediction architectures. Secondly, by utilising automatic differentiation in existing backpropagation frameworks, we explicitly optimise networks for risk-adjusted performance metrics, i.e. the Sharpe ratio \cite{SharpeRatio}, improving the risk profile of the signal on the whole. Lastly, retaining a consistent framework with other momentum strategies also allows us to retain desirable attributes from previous works -- specifically volatility scaling, which plays a critical role in the positive performance of time series momentum strategies \cite{VolTargeting}. This consistency also helps when making comparisons to existing methods, and facilitates the interpretation of different components of the overall signal by practitioners.

\section{Related Works}
\subsection{Classical Momentum Strategies}
Momentum strategies are traditionally divided into two categories -- namely (multivariate) cross sectional momentum \cite{CrossSectionalMomentum, XSectionalDeepMomentum} and (univariate) time series momentum \cite{TimeSeriesMomentum, DemystifyingTimeSeriesMomentum}. Cross sectional momentum strategies focus on the relative performance of securities against each other, buying relative winners and selling relative losers. By ranking a universe of stocks based on their past return and trading the top decile against the bottom decile, \cite{CrossSectionalMomentum} find that securities that recently outperformed their peers over the past 3 to 12 months continue to outperform on average over the next month. The performance of cross sectional momentum has also been shown to be stable across time \cite{CrossSectionalMomentum2}, and across a variety of markets and asset classes  \cite{AHLMomentum}. 

Time series momentum extends the idea to focus on an asset's own past returns, building portfolios comprising all securities under consideration. This was initially proposed by \cite{TimeSeriesMomentum}, who describe a concrete strategy which uses volatility scaling and trades positions based on the sign of returns over the past year -- demonstrating profitability across 58 different liquid instruments \textit{individually} over 25 years of data. Since then, numerous trading rules have been proposed -- with various trend estimation techniques and methods map them to traded positions. For instance, \cite{TrendFilteringLyxor} documents a wide range of linear and non-linear filters to measure trends and a statistic to test for its significance -- although methods to size positions with these estimates are not directly discussed. \cite{DemystifyingTimeSeriesMomentum} adopt a similar approach to \cite{TimeSeriesMomentum}, regressing the log price over the past 12 months against time and using the regression coefficient t-statistics to determine the direction of the traded position. While Sharpe ratios were comparable between the two, t-statistic based trend estimation led to a $66\%$ reduction in portfolio turnover and consequently trading costs. More sophisticated trading rules are proposed in \cite{AHLMomentum} and \cite{CurrencyMomentum}, taking volatility-normalised moving average convergence divergence (MACD) indicators as inputs. Despite the diversity of options, few comparisons have been made between the trading rules themselves, offering little clear evidence or intuitive reasoning to favour one rule over the next. We hence propose the use of deep neural networks to generate these rules directly, avoiding the need for explicit specification. Training them based on risk-adjusted performance metrics, the networks hence learn optimal training rules directly from the data itself. 

%Volatility scaling has been shown. Similar in principle to risk parity strategies, the . This has been 

\subsection{Deep Learning in Finance}
\label{sec:dl_in_finance}
Machine learning has long been used for financial time series prediction, with recent deep learning applications studying mid-price prediction using daily data \cite{ConveNetEnsembles}, or using limit order book data in a high frequency trading setting \cite{RamaContUniversality, DeepLOB, BDLOB}. While a variety of CNN and RNN models have been proposed, they typically frame the forecasting task as a classification problem, demonstrating the improved accuracy of their method in predicting the direction of the next price movement. Trading rules are then manually defined in relation to class probabilities -- either by using thresholds on classification probabilities to determine when to initiate positions \cite{ConveNetEnsembles}, or incorporating these thresholds into the classification problem itself by dividing price movements into buy, hold and sell classes depending on magnitude \cite{DeepLOB, BDLOB}. In addition to restricting the universe of strategies to those which rely on high accuracy, further gains might be made by learning trading rules directly from the data and removing the need for manual specification -- both of which are addressed in our proposed method.

Deep learning regression methods have also been considered in cross-sectional strategies \cite{EmpiricalAssetPricingViaML, XSectionalDeepMomentum}, ranking assets on the basis of expected returns over the next time period. Using a variety of linear, tree-based and neural network models \cite{EmpiricalAssetPricingViaML} demonstrate the outperformance of non-linear methods, with deep neural networks -- specifically 3-layer multilayer perceptrons (MLPs) -- having the best out-of-sample predictive $R^2$. Machine learning portfolios were then built by ranking stocks on a monthly basis using model predictions, with the best strategy coming from a 4-layer MLP that trades the top decile against the bottom decile of predictions. In other works, \cite{XSectionalDeepMomentum} adopt a similar approach using autoencoder and denoising autoencoder architectures, incorporating volatility scaling into their model as well.  While the results with basic deep neural networks are promising, they do not consider more modern architectures for time series prediction, such as the LSTM \cite{lstm} and WaveNet \cite{WaveNet} architectures which we evaluate for the DMN. Moreover, to the best of our knowledge, our paper is the first to consider the use of deep learning within the context of time series momentum strategies -- opening up possibilities in an alternate class of signals. 

Popularised by success of DeepMind's AlphaGo Zero \cite{AlphaGoZero}, deep reinforcement learning (RL) has also gained much attention in recent times -- prized for its ability to recommend path-dependent actions in dynamic environments. RL is particularly of interest within the context of optimal execution and automated hedging \cite{DynamicHedging, DeepHedging} for example, where actions taken can have an impact on future states of the world (e.g. market impact). However, deep RL methods generally require a realistic simulation environment (for Q-learning or policy gradient methods), or model of the world (for model-based RL) to provide feedback to agents during training -- both of which are difficult to obtain in practice.

\section{Strategy Definition}
Adopting the terminology of \cite{DemystifyingTimeSeriesMomentum}, the combined returns of a time series momentum (TSMOM) strategy can be expressed as below -- characterised by a trading rule or signal $X_t \in [-1, 1]$:

\begin{equation}
\label{eqn:tsmom}
r_{t,t+1}^{TSMOM} = \frac{1}{N_t} \sum_{i=1}^{N_t} X_t^{(i)}~\frac{\sigma_{\mathrm{tgt}}}{\sigma_t^{(i)}}~r_{t,t+1}^{(i)}.
\end{equation}

Here $r_{t, t+1}^{TSMOM}$ is the realised return of the strategy from day $t$ to $t+1$, $N_t$ is the number of included assets at $t$, and $r_{t, t+1}^{(i)}$ is the one-day return of asset $i$. We set the annualised volatility target $\sigma_{\mathrm{tgt}}$ to be $15\%$ and scale asset returns with an ex-ante volatility estimate $\sigma_t^{(i)}$ -- computed using an exponentially weighted moving standard deviation with a 60-day span on $r_{t, t+1}^{(i)}$. 

\subsection{Standard Trading Rules}
\label{sec:standard_trading_rules}
In traditional financial time series momentum strategies, the construction of a trading signal $X_t$ is typically divided into two steps: 1) estimating future trends based on past information, and 2) computing the actual positions to hold. We illustrate this in this section using two examples from the academic literature \cite{TimeSeriesMomentum,AHLMomentum}, which we also include as benchmarks into our tests. \\

\subsubsection*{Moskowitz et al. 2012 \cite{TimeSeriesMomentum}} In their original paper on time series momentum, a simple trading rule is adopted as below:
\begin{align}
&\text{\textbf{Trend Estimation: }}&  &Y_t^{(i)} = r_{t-252, t}^{(i)} \\
&\text{\textbf{Position Sizing:}} & & X_t^{(i)} = \sgn(Y_t^{(i)})
\label{eqn:moskowitz_sizing}
\end{align}
 
This broadly uses the past year's returns as a trend estimate for the next time step - taking a maximum long position when the expected trend is positive (i.e. $\sgn(r_{t-252,t}^{(i)})$) and a maximum short position when negative. \\

\subsubsection*{Baz et al. 2015 \cite{AHLMomentum}} In practice, more sophisticated methods can be used to compute $Y_t^{(i)}$ and $X_t^{(i)}$ -- such as the model of \cite{AHLMomentum} described below: 

\begin{align}
\label{eqn:ahl_macd}
&\text{\textbf{Trend Estimation: }}  && Y_t^{(i)} =  \frac{q_t^{(i)} }{\mathrm{std}(z_{t-252:t}^{(i)})}
\end{align}
\begin{align}
&q_t^{(i)} =  \mathrm{MACD}(i,t, S, L) ~/~  \mathrm{std}(p_{t-63:t}^{(i)}) && \\
& \mathrm{MACD}(i,t, S, L) = m(i, S) - m(i, L). &&
\end{align}
Here $\mathrm{std}(p_{t-63:t} ^{(i)})$ is the 63-day rolling standard deviation of asset $i$ prices $p^{(i)}_{t-63: t} = [p^{(i)}_{t-63}, \dots, p^{(i)}_t]$, $m(i, S)$ is the exponentially weighted moving average of asset $i$ prices with a time-scale $S$ that translates into a half-life of $HL=\log(0.5)/\log(1-\frac{1}{S})$. The moving average crossover divergence (MACD) signal is defined in relation to a short and a long time-scale $S$ and $L$ respectively. 

\begin{figure}[h]
\caption{Position Sizing Function $\phi(y)$}
\label{fig:SizingFunction}
\centering
\includegraphics[width=1\linewidth]{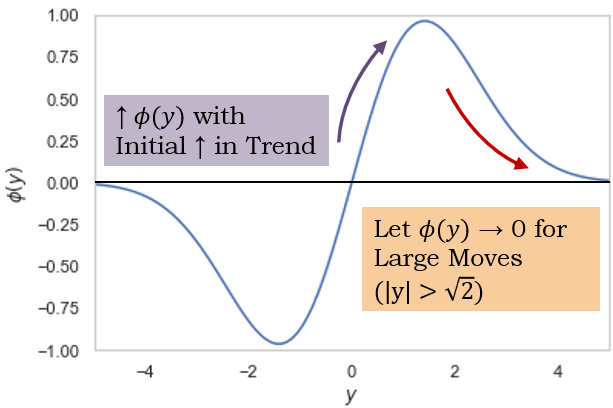}
\end{figure}

The volatility-normalised MACD signal hence measures the strength of the trend, which is then translated in to a position size as below:

\begin{align}
&\text{\textbf{Position Sizing: }} && X_t^{(i)} = \phi(Y_t^{(i)}),
\end{align}
where $\phi(y) = \frac{y\exp(\frac{-y^2}{4})}{0.89}$. Plotting $\phi(y)$ in Exhibit \ref{fig:SizingFunction}, we can see that positions are increased until $|Y_t^{(i)}|= \sqrt{2} \approx 1.41$, before decreasing back to zero for larger moves. This allows the signal to reduces positions in instances where assets are overbought or oversold -- defined to be when $|q_t^{(i)}|$ is observed to be larger than 1.41 times its past year's standard deviation.

Increasing the complexity even further, multiple signals with different times-scales can also be averaged to give a final position:

\begin{equation}
\tilde{Y}_t^{(i)} = \sum_{k=1}^3 Y_t^{(i)} (S_k, L_k),
\end{equation}
where $Y_t^{(i)} (S_k, L_k)$ is as per Equation \eqref{eqn:ahl_macd} with explicitly defined short and long time-scales -- using $S_k \in \{8, 16, 32\}$ and $L_k \in \{24, 48, 96\}$ as defined in \cite{AHLMomentum}.

\subsection{Machine Learning Extensions}
\label{sec:ml_extensions}
As can be seen from Section \ref{sec:standard_trading_rules}, many explicit design decisions are required to define a sophisticated time series momentum strategy. We hence start by considering how machine learning methods can be used to learn these relationships directly from data -- alleviating the need for manual specification.\\

\subsubsection*{Standard Supervised Learning} In line with numerous previous works (see Section \ref{sec:dl_in_finance}), we can cast trend estimation as a standard regression or binary classification problem, with outputs:
\begin{align}
&\text{\textbf{Trend Estimation: }}&  & Y_t^{(i)} = f\left(\mathbf{u}_t^{(i)}; \pmb{\theta} \right),
\end{align}
where $f(\cdot)$ is the output of the machine learning model, which takes in a vector of input features $\mathbf{u}_t^{(i)}$ and model parameters  $\pmb{\theta}$ to generate predictions. Taking volatility-normalised returns as targets, the following mean-squared error and binary cross-entropy losses can be used for training:
\begin{eqnarray}
\mathcal{L}_{\mathrm{reg}} (\pmb{\theta}) &=& \frac{1}{M}   \sum_\Omega \left(Y_t^{(i)} - \frac{r_{t, t+1}^{(i)} }{\sigma_t^{(i)}}  \right)^2 \\
\mathcal{L}_{\mathrm{binary}} (\pmb{\theta}) &=& - \frac{1}{M}   \sum_\Omega   \bigg\{ \mathbb{I}~\log \left(Y_t^{(i)} \right) \nonumber \\ 
&&  + \left(1-\mathbb{I} \right)\log\left(1- Y_t^{(i)} \right)  \bigg\},
\end{eqnarray}
where $\Omega = \big\{ \big(Y_1^{(1)}, r_{1, 2}^{(1)} /\sigma_t^{(1)}\big), ~\dots~,$ $\big(Y_{T-1}^{(N)}, r_{T-1, T}^{(N)} / \sigma_{T-1}^{(N)}\big) \big\}$ is the set of all $M$ possible prediction and target tuples across all $N$ assets and $T$ time steps. For the binary classification case, $\mathbb{I}$ is the indicator function $\mathbb{I}\big(r_{t, t+1}^{(i)} / \sigma_t^{(i)} > 0\big)$ -- making $Y_t^{(i)}$ the estimated probability of a positive return. 

This still leaves us to specify how trend estimates map to positions, and we do so using a similar form to Equation \ref{eqn:moskowitz_sizing}: \\

\textbf{Position Sizing:}
\begin{align}
&\text{   Regression }&  & X_t^{(i)} = \sgn(Y_t^{(i)}) &&\\
&\text{   Classification }&  & X_t^{(i)} = \sgn(Y_t^{(i)} - 0.5) &&
\end{align}

As such, we take a maximum long position when the expected returns are positive in the regression case, or when the probability of a positive return is greater than 0.5 in the classification case. \\

\subsubsection*{Direct Outputs} An alternative approach is to use machine learning models to generate positions directly -- simultaneously learning both trend estimation and position sizing in the same function, i.e.:
\begin{align}
&\text{\textbf{Direct Outputs: }}&  & X_t^{(i)} = f\left(\mathbf{u}_t^{(i)}; \pmb{\theta} \right).
\end{align}
Given the lack of direct information on the optimal positions to hold at each step -- which is required to produce labels for standard regression and classification models -- calibration would hence need to be performed by directly optimising performance metrics. Specifically, we focus on optimising the average return and the annualised Sharpe ratio via the loss functions below:
\begin{eqnarray}
 \!\! \mathcal{L}_{\mathrm{returns}} (\pmb{\theta}) \!\! \!\! &= &  -\mu_R \nonumber \\
&=& - \frac{1}{M} \sum_\Omega R(i, t) \nonumber \\
&=&- \frac{1}{M} \sum_\Omega X_t^{(i)}~\frac{\sigma_{\mathrm{tgt}}}{\sigma_t^{(i)}}~r_{t,t+1}^{(i)} \\
\!\!  \mathcal{L}_{\mathrm{sharpe}} (\pmb{\theta})\!\! \!\!  &=&\!\!\!  - \frac{ \mu_R \times \sqrt{252} }{\sqrt{ \left(\sum_\Omega R(i, t)^2 \right)/ M- \mu_R^2}}
\end{eqnarray}
where $\mu_R$ is the average return over $\Omega$, and $R(i,t)$ is the return captured by the trading rule for asset $i$ at time $t$.

\section{Deep Momentum Networks}
In this section, we examine a variety of architectures that can be used in Deep Momentum Networks -- all of which can be easily reconfigured to generate the predictions described in Section \ref{sec:ml_extensions}. This is achieved by implementing the models using the \texttt{Keras} API in \texttt{Tensorflow} \cite{tensorflow}, where output activation functions can be flexibly interchanged to generate the predictions of different types (e.g. expected returns, binary probabilities, or direct positions). Arbitrary loss functions can also be defined for direct outputs, with gradients for backpropagation being easily computed using the built-in libraries for automatic differentiation.

\subsection{Network Architectures}
\paragraph*{Lasso Regression} In the simplest case, a standard linear model could be used to generate predictions as below:
\begin{equation}
Z_t^{(i)} = g\left( \mathbf{w}^T \mathbf{u}_{t-\tau:t}^{(i)} + b \right),
\end{equation}
where $Z_t^{(i)} \in \left\{X_t^{(i)}, Y_t^{(i)}  \right\}$ depending on the prediction task,  $\mathbf{w}$ is a weight vector for the linear model, and $b$ is a bias term. Here $g(\cdot)$ is a activation function which depends on the specific prediction type -- linear for standard regression, sigmoid for binary classification, and tanh-function for direct outputs. 

Additional regularisation is also provided during training by augmenting the various loss functions to include an additional $L_1$ regulariser as below:

\begin{equation}
\tilde{\mathcal{L}}(\pmb{\theta}) = \mathcal{L}(\pmb{\theta}) + \alpha || \mathbf{w} ||_1,
\end{equation}
where $\mathcal{L}(\pmb{\theta})$ corresponds to one of the loss functions described in Section \ref{sec:ml_extensions}, $|| \mathbf{w} ||_1$ is the $L_1$ norm of $\mathbf{w}$, and $\alpha$ is a constant term which we treat as an additional hyperparameter. To incorporate recent history into predictions as well, we concatenate inputs over the past $\tau$-days into a single input vector -- i.e. $\mathbf{u}_{t-\tau:t}^{(i)} = [\mathbf{u}_{t-\tau}^{(i) ~T}, \dots, \mathbf{u}_{t}^{(i)~T}]^T$. This was fixed to be $\tau=5$ days for tests in Section \ref{sec:tests}. \\

\paragraph*{Multilayer Perceptron (MLP)} Increasing the degree of model complexity slightly, a 2-layer neural network can be used to incorporated non-linear effects:
\begin{align}
\mathbf{h}_t^{(i)} &= \tanh \left( \mathbf{W}_h \mathbf{u}_{t-\tau:t}^{(i)} + \mathbf{b}_h \right)\\
Z_t^{(i)} &= g\left( \mathbf{W}_z \mathbf{h}_t^{(i)} + \mathbf{b}_z \right),
\end{align}
where $\mathbf{h}_t^{(i)}$ is the hidden state of the MLP using an internal tanh activation function, $\tanh(\cdot)$, and $\mathbf{W}_.$ and $\mathbf{b}_.$ are layer weight matrices and biases respectively.\\

\paragraph*{WaveNet} More modern techniques such as convolutional neural networks (CNNs) have been used in the domain of time series prediction -- particularly in the form of autoregressive architectures e.g. \cite{AutoregressiveCNNs}. These typically take the form of 1D causal convolutions, sliding convolutional filters across time to extract useful representations which are then aggregated in higher layers of the network. To increase the size of the receptive field -- or the length of history fed into the CNN -- dilated CNNs such as WaveNet \cite{WaveNet} have been proposed, which skip over inputs at intermediate levels with a predetermined dilation rate. This allows it to effectively increase the amount of historical information used by the CNN without a large increase in computational cost. Let us consider a dilated convolutional layer with residual connections take the form below: 
\begin{align}
\psi(\mathbf{u}) =& \underbrace{\tanh(\mathbf{W}\mathbf{u}) \odot \sigma(\mathbf{V}\mathbf{u})}_\text{Gated Activation} \nonumber \\
& + \underbrace{\mathbf{A}\mathbf{u} + \mathbf{b}}_\text{Skip Connection}.
\end{align}
Here $\mathbf{W}$ and $\mathbf{V}$ are weight matrices associated with the gated activation function, and $\mathbf{A}$ and $\mathbf{b}$ are the weights and biases used to transform the $\mathbf{u}$ to match dimensionality of the layer outputs for the skip connection. The equations for WaveNet architecture used in our investigations can then be expressed as:
\begin{eqnarray}
\mathbf{s}_{\mathrm{weekly}}^{(i)}(t) &=& \psi(\mathbf{u}_{t-5:t}^{(i)})  \label{eqn:cnn_input} \\ 
%\end{equation}
%\begin{equation}
\mathbf{s}_{\mathrm{monthly}}^{(i)}(t) &=& \psi\left( \begin{bmatrix} \mathbf{s}_{\mathrm{weekly}}^{(i)}(t) \\ \mathbf{s}_{\mathrm{weekly}}^{(i)}(t-5) \\ \mathbf{s}_{\mathrm{weekly}}^{(i)}(t-10) \\ \mathbf{s}_{\mathrm{weekly}}^{(i)}(t-15) \end{bmatrix}\right) \\
%\end{equation}
%\begin{equation}
\mathbf{s}_{\mathrm{quarterly}}^{(i)}(t)&=& \psi\left( \begin{bmatrix} \mathbf{s}_{\mathrm{monthly}}^{(i)}(t) \\ \mathbf{s}_{\mathrm{monthly}}^{(i)}(t-21) \\ \mathbf{s}_{\mathrm{monthly}}^{(i)}(t-42) \end{bmatrix}\right).\label{eqn:cnn_quarterly}
\end{eqnarray}
Here each intermediate layer $\mathbf{s}_{.}^{(i)}(t)$ aggregates representations at weekly, monthly and quarterly frequencies respectively. Intermediate layers are then concatenated at each layer before passing through a 2-layer MLP to generate outputs, i.e.:
\begin{align}
\mathbf{s}_t^{(i)} &= \begin{bmatrix} \mathbf{s}_{\mathrm{weekly}}^{(i)}(t) \\ \mathbf{s}_{\mathrm{monthly}}^{(i)}(t) \\ \mathbf{s}_{\mathrm{quarterly}}^{(i)}(t) \end{bmatrix} \\
\label{eqn:cnn_mlp}
\mathbf{h}_t^{(i)} &= \tanh(\mathbf{W}_h \mathbf{s}_t^{(i)} + \mathbf{b}_h ) \\
Z_t^{(i)} &= g\left( \mathbf{W}_z \mathbf{h}_t^{(i)} + \mathbf{b}_z \right).
\end{align}
State sizes for each intermediate layers $\mathbf{s}_{\mathrm{weekly}}^{(i)}(t)$, $\mathbf{s}_{\mathrm{monthly}}^{(i)}(t)$, $\mathbf{s}_{\mathrm{quarterly}}^{(i)}(t) $ and the MLP hidden state $\mathbf{h}_t^{(i)}$ are fixed to be the same, allowing us to use a single hyperparameter to define the architecture. To independently evaluate the performance of CNN and RNN architectures, the above also excludes the LSTM block (i.e. the context stack) described in \cite{WaveNet}, focusing purely on the merits of the dilated CNN model.    \\

\paragraph*{Long Short-term Memory (LSTM)} Traditionally used in sequence prediction for natural language processing, recurrent neural networks -- specifically long short-term memory (LSTM) architectures \cite{lstm} -- have been increasing used in time series prediction tasks. The equations for the LSTM in our model are provided below:
\begin{align}
&\mathbf{f}_t^{(i)} &&= \sigma(\mathbf{W}_{f} \mathbf{u}_t^{(i)}  + \mathbf{V}_{f} \mathbf{h}_{t-1}^{(i)} + \mathbf{b}_f ) \\
&\mathbf{i}_t^{(i)} &&= \sigma(\mathbf{W}_{i} \mathbf{u}_t^{(i)}  + \mathbf{V}_{i} \mathbf{h}_{t-1}^{(i)} + \mathbf{b}_i )\\
&\mathbf{o}_t^{(i)} &&= \sigma(\mathbf{W}_{o} \mathbf{u}_t^{(i)}  + \mathbf{V}_{o} \mathbf{h}_{t-1}^{(i)} + \mathbf{b}_o )\\
&\mathbf{c}_t^{(i)} &&= \mathbf{f}_t^{(i)} \odot \mathbf{c}_{t-1}^{(i)}  \nonumber \\
&&& + \mathbf{i}_t^{(i)} \odot \tanh( \mathbf{W}_{c} \mathbf{u}_t^{(i)}  + \mathbf{V}_{c} \mathbf{h}_{t-1}^{(i)} + \mathbf{b}_c  )\\
&\mathbf{h}_t^{(i)} &&= \mathbf{o}_t^{(i)} \odot \tanh(\mathbf{c}_t^{(i)}) \\
&Z_t^{(i)} &&= g \left( \mathbf{W}_z \mathbf{h}_t^{(i)} + \mathbf{b}_z \right),
\end{align}
where $\odot$ is the Hadamard (element-wise) product, $\sigma(.)$ is the sigmoid activation function, $\mathbf{W}_{.}$ and $\mathbf{V}_{.}$ are weight matrices for the different layers, $\mathbf{f}_t^{(i)}, \mathbf{i}_t^{(i)}, \mathbf{o}_t^{(i)}$ correspond to the forget, input and output gates respectively, $\mathbf{c}_t^{(i)}$ is the cell state, and $\mathbf{h}_t^{(i)}$ is the hidden state of the LSTM. From these equations, we can see that the LSTM uses the cell state as a compact summary of past information, controlling memory retention with the forget gate and incorporating new information via the input gate. As such, the LSTM is able to learn representations of long-term relationships relevant to the prediction task -- sequentially updating its internal memory states with new observations at each step.

\subsection{Training Details}
\label{sec:training}
Model calibration was undertaken using minibatch stochastic gradient descent with the \texttt{Adam} optimiser \cite{ADAM}, based on the loss functions defined in Section \ref{sec:ml_extensions}. Backpropagation was performed up to a maximum of 100 training epochs using $90\%$ of a given block of training data, and the most recent $10 \%$ retained as a validation dataset. Validation data is then used to determine convergence -- with early stopping triggered when the validation loss has not improved for 25 epochs -- and to identify the optimal model across hyperparameter settings. Hyperparameter optimisation was conducted using 50 iterations of random search, with full details provided in Appendix \ref{apdx:hyperparam}.  For additional information on the deep neural network calibration, please refer to \cite{DeepLearningBook}.

Dropout regularisation \cite{dropout} was a key feature to avoid overfitting in the neural network models -- with dropout rates included as hyperparameters during training. This was applied to the inputs and hidden state for the MLP, as well as the inputs, Equation \eqref{eqn:cnn_input}, and outputs, Equation \eqref{eqn:cnn_mlp}, of the convolutional layers in the WaveNet architecture. For the LSTM, we adopted the same dropout masks as in \cite{variationaldropout} -- applying dropout to the RNN inputs, recurrent states and outputs. \\

\section{Performance Evaluation}
\label{sec:tests}
\subsection{Overview of Dataset}
The predictive performance of the different architectures was evaluated via a backtest using 88 ratio-adjusted continuous futures contracts downloaded from the Pinnacle Data Corp CLC Database \cite{PinnacleData}. These contracts spanned across a variety of asset classes -- including commodities, fixed income and currency futures -- and contained prices from 1990 to 2015. A full breakdown of the dataset can be found in Appendix \ref{apdx:data}.

\subsection{Backtest Description}
\label{sec:backtest}
Throughout our backtest, the models were recalibrated from scratch every 5 years -- re-running the entire hyperparameter optimisation procedure using all data available up to the recalibration point. Model weights were then fixed for signals generated over the next 5 year period, ensuring that tests were performed out-of-sample. 

For the Deep Momentum Networks, we incorporate a series of useful features adopted by standard time series momentum strategies in Section \ref{sec:standard_trading_rules} to generate predictions at each step: 

\begin{enumerate}
\item \textit{Normalised Returns} -- Returns over the past day, 1-month, 3-month, 6-month and 1-year periods are used, normalised by a measure of daily volatility scaled to an appropriate time scale. For instance, normalised annual returns were taken to be $r_{t-252, t}^{(i)} / (\sigma_t^{(i)}  \sqrt{252}$).
\item \textit{MACD Indicators} -- We also include the MACD indicators -- i.e. trend estimates $Y_t^{(i)}$ -- as in Equation \eqref{eqn:ahl_macd}, using the same short time-scales $S_k \in \{8, 16, 32\}$ and long time-scales $L_k \in \{24, 48, 96\}$.  
\end{enumerate}

For comparisons against traditional time series momentum strategies, we also incorporate the following reference benchmarks:

\begin{enumerate}
\item Long Only with Volatility Scaling  $(X_t^{(i)} = 1)$
\item Sgn(Returns) -- Moskowitz et al. 2012 \cite{TimeSeriesMomentum} 
\item MACD Signal -- Baz et al. 2015 \cite{AHLMomentum} 
\end{enumerate}

Finally, performance was judged based on the following metrics:
\begin{enumerate}
\item \textit{Profitability} --  Expected returns ($\mathbb{E}[\text{Returns}]$) and the percentage of positive returns observed across the test period.
\item \textit{Risk} -- Daily volatility (Vol.), downside deviation and the maximum drawdown (MDD) of the overall portfolio.
\item \textit{Performance Ratios} -- Risk adjusted performance was measured by the Sharpe ratio $\left(\frac{\mathbb{E}[\text{Returns}]}{\text{Vol.}}\right)$, Sortino ratio $\left(\frac{\mathbb{E}[\text{Returns}]}{\text{Downside Deviation}}\right)$ and Calmar ratio $\left(\frac{\mathbb{E}[\text{Returns}]}{\text{MDD}}\right)$, as well as the average profit over the average loss $\left(\frac{\text{Ave. P}}{\text{Ave. L}} \right)$.
\end{enumerate}

\subsection{Results and Discussion}
\begin{figure*}[htbp]
\caption{Cumulative Returns - Rescaled to Target Volatility}
\label{fig:cumulative_returns}
\centering
\begin{subfigure}[]{0.475\linewidth}
\fbox{\includegraphics[width=1\linewidth]{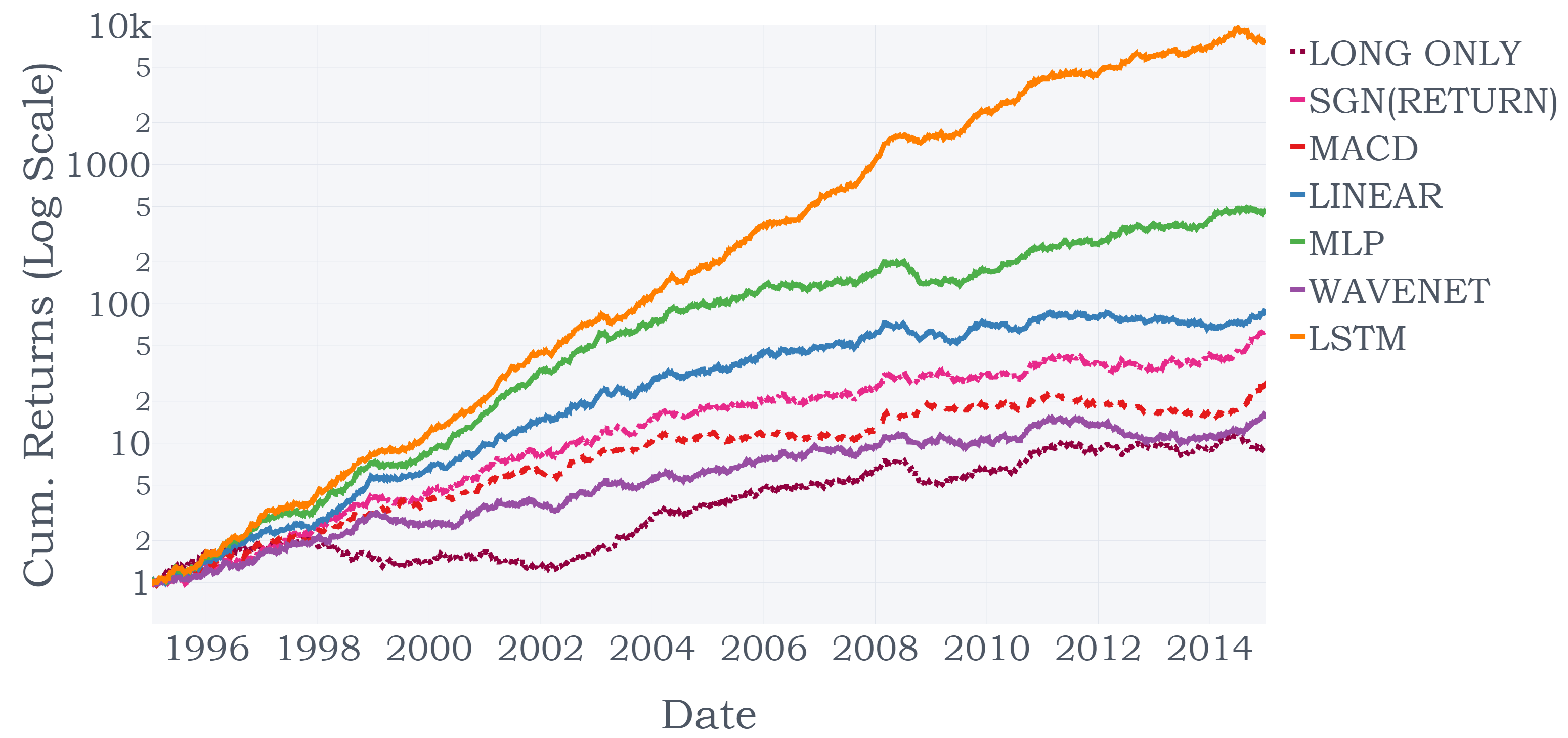}}
\caption{Sharpe Ratio}
\end{subfigure}\hfill
\begin{subfigure}[]{0.475\linewidth}
\fbox{\includegraphics[width=1\linewidth]{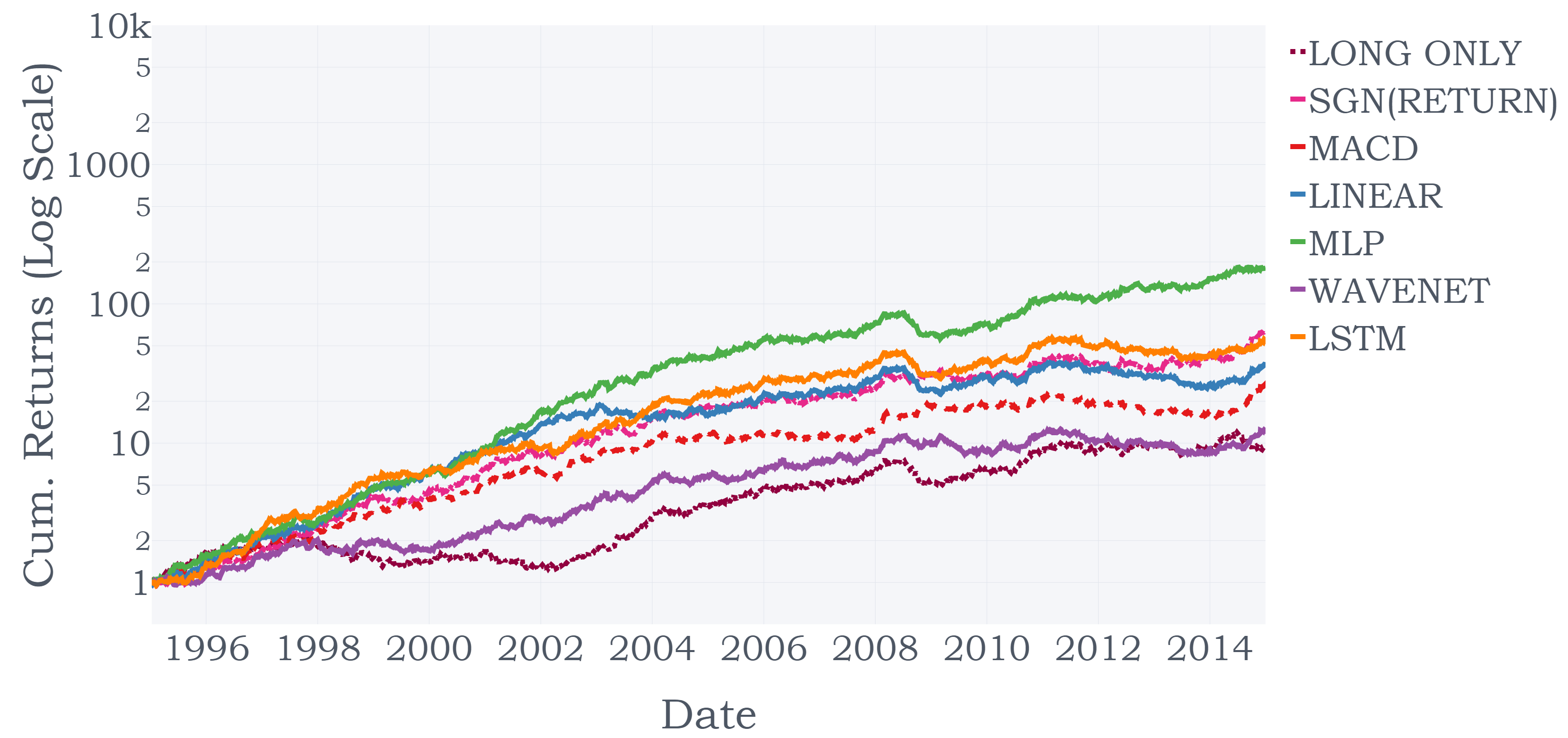}}
\caption{Average Returns}
\end{subfigure}\\
\begin{subfigure}[]{0.475\linewidth}
\fbox{\includegraphics[width=1\linewidth]{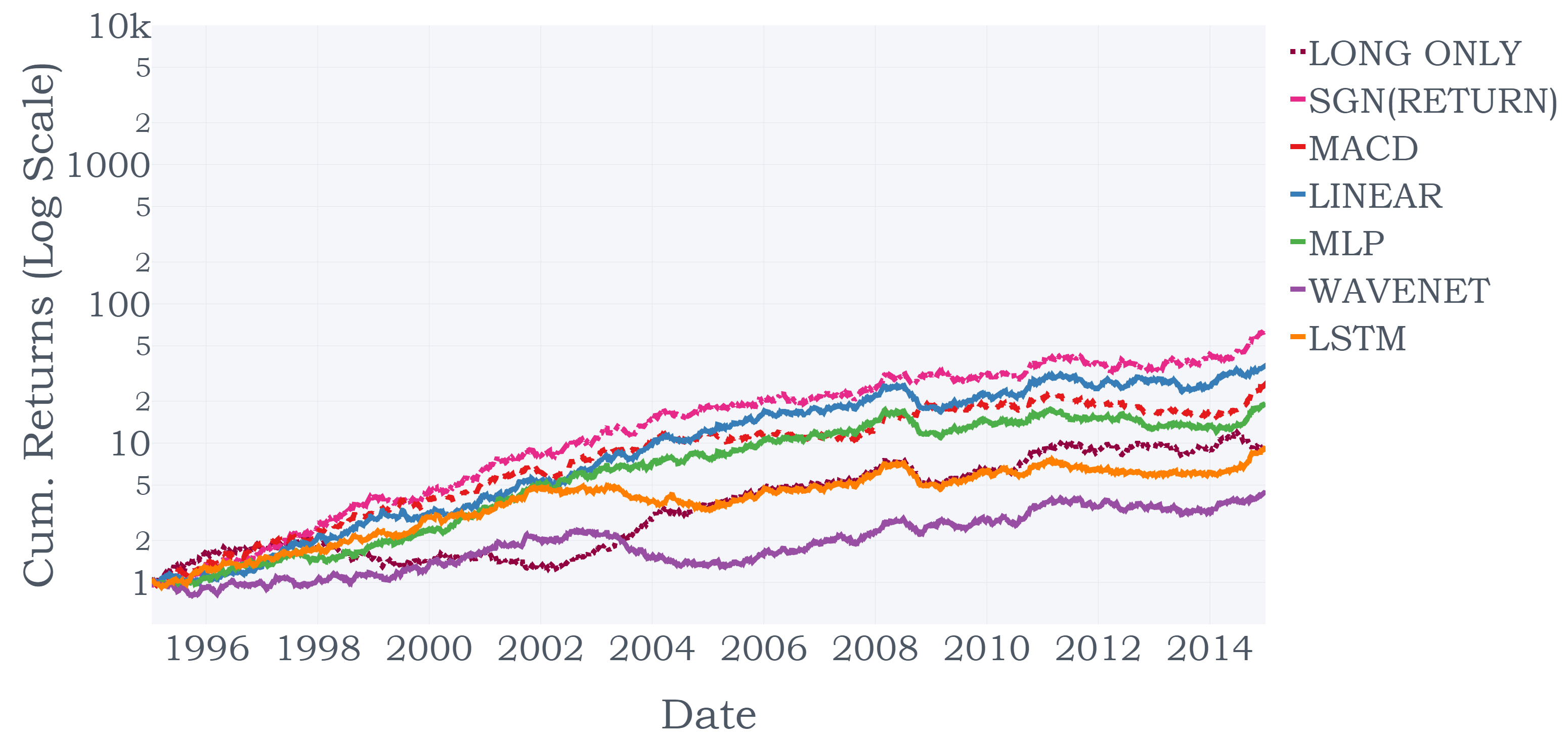}}
\caption{MSE}
\end{subfigure}\hfill
\begin{subfigure}[]{0.475\linewidth}
\fbox{\includegraphics[width=1\linewidth]{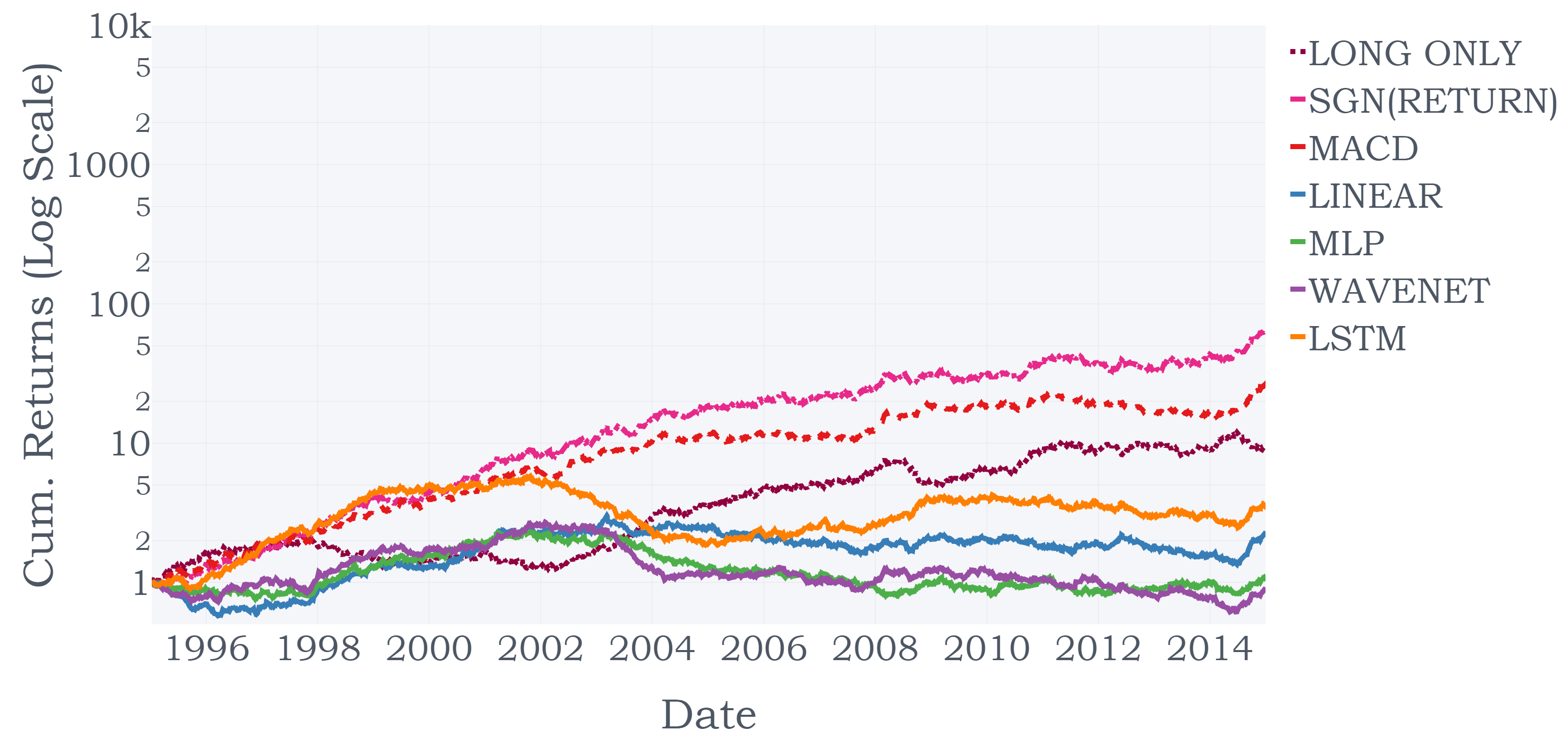}}
\caption{Binary}
\end{subfigure}
\end{figure*}

Aggregating the out-of-sample predictions from 1995 to 2015, we compute performance metrics for both the strategy returns based on Equation \eqref{eqn:tsmom} (Exhibit \ref{table:performanceraw}), as well as that for portfolios with an additional layer of volatility scaling -- which brings overall strategy returns to match the $15\%$ volatility target (Exhibit \ref{table:performancerescaled}). Given the large differences in returns volatility seen in Table \ref{table:performanceraw}, this rescaling also helps to facilitate comparisons between the cumulative returns of different strategies -- which are plotted for various loss functions in Exhibit \ref{fig:cumulative_returns}. We note that strategy returns in this section are computed in the absence of transaction costs, allowing us to focus on the raw predictive ability of the models themselves. The impact of transaction costs is explored further in Section \ref{sec:costs}, where we undertake a deeper analysis of signal turnover. More detailed results can also be found in Appendix \ref{apdx:extra_results}, which echo the findings below.

Focusing on the raw signal outputs, the Sharpe ratio-optimised LSTM outperforms all benchmarks as expected, improving the best neural network model (Sharpe-optimised MLP) by $44 \%$ and the best reference benchmark (Sgn(Returns)) by more than two times. In conjunction with Sharpe ratio improvements to both the linear and MLP models, this highlights the benefits of using models which capture non-linear relationships, and have access to more time history via an internal memory state. Additional model complexity, however, does not necessarily lead to better predictive performance, as demonstrated by the underperformance of WaveNet compared to both the reference benchmarks and simple linear models. Part of this can be attributed to the difficulties in tuning models with multiple design parameters - for instance, better results could possibly achieved by using alternative dilation rates, number of convolutional layers, and hidden state sizes in Equations \eqref{eqn:cnn_input} to \eqref{eqn:cnn_quarterly} for the WaveNet. In contrast, only a single design parameter is sufficient to specify the hidden state size in both the MLP and LSTM models. Analysing the relative performance within each model class, we can see that models which directly generate positions perform the best -- demonstrating the benefits of simultaneous learning both trend estimation and position sizing functions. In addition, with the exception of a slight decrease in the MLP, Sharpe-optimised models outperform returns-optimised ones, with standard regression and classification benchmarks taking third and fourth place respectively. 

\begin{table*}[tbhp]
\centering
\caption{Performance Metrics -- Raw Signal Outputs}
\label{table:performanceraw}
\begin{tabular}{llllllllll}
\hline
\textbf{}                 & \textbf{E[Return]} & \textbf{Vol.}   & \textbf{\begin{tabular}[c]{@{}l@{}}Downside\\     Deviation\end{tabular}} & \textbf{MDD}    & \textbf{Sharpe} & \textbf{Sortino} & \textbf{Calmar} & \textbf{\begin{tabular}[c]{@{}l@{}}\% of $+$ve \\     Returns\end{tabular}} & \textbf{$\mathbf{\frac{\text{Ave.  P}}{\text{Ave. L}}}$} \\ \midrule
{\ul \textbf{Reference}} &                    &                 &                                                                           &                 &                 &                  &                 &                                                                             &                                                          \\
Long Only                 & 0.039              & 0.052           & 0.035                                                                     & 0.167           & 0.738           & 1.086            & 0.230           & 53.8\%                                                                      & 0.970                                                    \\
Sgn(Returns)             & 0.054              & 0.046           & 0.032                                                                     & 0.083           & 1.192           & 1.708            & 0.653           & 54.8\%                                                                      & 1.011                                                    \\
MACD                      & 0.030              & 0.031           & 0.022                                                                     & 0.081           & 0.976           & 1.356            & 0.371           & 53.9\%                                                                      & 1.015                                                    \\ \midrule
{\ul \textbf{Linear}}     &                    &                 &                                                                           &                 &                 &                  &                 &                                                                             &                                                          \\
Sharpe                    & 0.041              & 0.038           & 0.028                                                                     & 0.119           & 1.094           & 1.462            & 0.348           & 54.9\%                                                                      & 0.997                                                    \\
Ave. Returns              & 0.047              & 0.045           & 0.031                                                                     & 0.164           & 1.048           & 1.500            & 0.287           & 53.9\%                                                                      & 1.022                                                    \\
MSE                       & 0.049              & 0.047           & 0.032                                                                     & 0.164           & 1.038           & 1.522            & 0.298           & 54.3\%                                                                      & 1.000                                                    \\
Binary                    & 0.013              & 0.044           & 0.030                                                                     & 0.167           & 0.295           & 0.433            & 0.078           & 50.6\%                                                                      & 1.028                                                    \\ \midrule
{\ul \textbf{MLP}}        &                    &                 &                                                                           &                 &                 &                  &                 &                                                                             &                                                          \\
Sharpe                    & 0.044              & 0.031           & 0.025                                                                     & 0.154           & 1.383           & 1.731            & 0.283           & 56.0\%                                                                      & 1.024                                                    \\
Ave. Returns              & \textbf{0.064*}    & 0.043           & 0.030                                                                     & 0.161           & 1.492           & 2.123            & 0.399           & 55.6\%                                                                      & 1.031                                                    \\
MSE                       & 0.039              & 0.046           & 0.032                                                                     & 0.166           & 0.844           & 1.224            & 0.232           & 52.7\%                                                                      & 1.035                                                    \\
Binary                    & 0.003              & 0.042           & 0.028                                                                     & 0.233           & 0.080           & 0.120            & 0.014           & 50.8\%                                                                      & 0.981                                                    \\ \midrule
{\ul \textbf{WaveNet}}    &                    &                 &                                                                           &                 &                 &                  &                 &                                                                             &                                                          \\
Sharpe                    & 0.030              & 0.035           & 0.026                                                                     & 0.101           & 0.854           & 1.167            & 0.299           & 53.5\%                                                                      & 1.008                                                    \\
Ave. Returns              & 0.032              & 0.040           & 0.028                                                                     & 0.113           & 0.788           & 1.145            & 0.281           & 53.8\%                                                                      & 0.980                                                    \\
MSE                       & 0.022              & 0.042           & 0.028                                                                     & 0.134           & 0.536           & 0.786            & 0.166           & 52.4\%                                                                      & 0.994                                                    \\
Binary                    & 0.000              & 0.043           & 0.029                                                                     & 0.313           & 0.011           & 0.016            & 0.001           & 50.2\%                                                                      & 0.995                                                    \\ \midrule
{\ul \textbf{LSTM}}       &                    &                 &                                                                           &                 &                 &                  &                 &                                                                             &                                                          \\
Sharpe                    & 0.045              & \textbf{0.016*} & \textbf{0.011*}                                                           & \textbf{0.021*} & \textbf{2.804*} & \textbf{3.993*}  & \textbf{2.177*} & \textbf{59.6\%*}                                                            & \textbf{1.102*}                                          \\
Ave. Returns              & 0.054              & 0.046           & 0.033                                                                     & 0.164           & 1.165           & 1.645            & 0.326           & 54.8\%                                                                      & 1.003                                                    \\
MSE                       & 0.031              & 0.046           & 0.032                                                                     & 0.163           & 0.669           & 0.959            & 0.189           & 52.8\%                                                                      & 1.003                                                    \\
Binary                    & 0.012              & 0.039           & 0.026                                                                     & 0.255           & 0.300           & 0.454            & 0.046           & 51.0\%                                                                      & 1.012                                                    \\ \hline
\end{tabular}
\vfill
\end{table*}

% \usepackage{booktabs}
% \usepackage[normalem]{ulem}
\begin{table*}[bthp]
\vfill
\centering
\caption{Performance Metrics -- Rescaled to Target Volatility}
\label{table:performancerescaled}
\begin{tabular}{@{}llllllllll@{}}
\hline
\textbf{}                 & \textbf{E{[}Return{]}} & \textbf{Vol.}   & \textbf{\begin{tabular}[c]{@{}l@{}}Downside\\     Deviation\end{tabular}} & \textbf{MDD}    & \textbf{Sharpe} & \textbf{Sortino} & \textbf{Calmar} & \textbf{\begin{tabular}[c]{@{}l@{}}\% of $+$ve \\     Returns\end{tabular}} & \textbf{\begin{tabular}[c]{@{}l@{}}$\mathbf{\frac{\text{Ave.  P}}{\text{Ave. L}}}$\end{tabular}} \\ \midrule
{\ul \textbf{Reference}} &                        &                 &                                                                      &                 &                 &                  &                 &                                                                             &                                                                                                     \\
Long Only                 & 0.117                  & 0.154           & 0.102                                                                & 0.431           & 0.759           & 1.141            & 0.271           & 53.8\%                                                                      & 0.973                                                                                               \\
Sgn(Returns)             & 0.215                  & 0.154           & 0.102                                                                & 0.264           & 1.392           & 2.108            & 0.815           & 54.8\%                                                                      & 1.041                                                                                               \\
MACD                      & 0.172                  & 0.155           & 0.106                                                                & 0.317           & 1.111           & 1.622            & 0.543           & 53.9\%                                                                      & 1.031                                                                                               \\ \midrule
{\ul \textbf{Linear}}     &                        &                 &                                                                      &                 &                 &                  &                 &                                                                             &                                                                                                     \\
Sharpe                    & 0.232                  & 0.155           & 0.103                                                                & 0.303           & 1.496           & 2.254            & 0.765           & 54.9\%                                                                      & 1.056                                                                                               \\
Ave. Returns              & 0.189                  & 0.154           & 0.100                                                                & 0.372           & 1.225           & 1.893            & 0.507           & 53.9\%                                                                      & 1.047                                                                                               \\
MSE                       & 0.186                  & 0.154           & \textbf{0.099*}                                                      & 0.365           & 1.211           & 1.889            & 0.509           & 54.3\%                                                                      & 1.025                                                                                               \\
Binary                    & 0.051                  & 0.155           & 0.103                                                                & 0.558           & 0.332           & 0.496            & 0.092           & 50.6\%                                                                      & 1.033                                                                                               \\ \midrule
{\ul \textbf{MLP}}        &                        &                 &                                                                      &                 &                 &                  &                 &                                                                             &                                                                                                     \\
Sharpe                    & 0.312                  & 0.154           & 0.102                                                                & 0.335           & 2.017           & 3.042            & 0.930           & 56.0\%                                                                      & 1.104                                                                                               \\ 
Ave. Returns              & 0.266                  & 0.154           & \textbf{0.099*}                                                      & 0.354           & 1.731           & 2.674            & 0.752           & 55.6\%                                                                      & 1.065                                                                                               \\
MSE                       & 0.156                  & 0.154           & \textbf{0.099*}                                                      & 0.371           & 1.017           & 1.582            & 0.422           & 52.7\%                                                                      & 1.062                                                                                               \\
Binary                    & 0.017                  & 0.154           & 0.102                                                                & 0.661           & 0.108           & 0.162            & 0.025           & 50.8\%                                                                      & 0.986                                                                                               \\ \midrule
{\ul \textbf{WaveNet}}    &                        &                 &                                                                      &                 &                 &                  &                 &                                                                             &                                                                                                     \\
Sharpe                    & 0.148                  & 0.155           & 0.103                                                                & 0.349           & 0.956           & 1.429            & 0.424           & 53.5\%                                                                      & 1.018                                                                                               \\
Ave. Returns              & 0.136                  & 0.154           & 0.101                                                                & 0.356           & 0.881           & 1.346            & 0.381           & 53.8\%                                                                      & 0.993                                                                                               \\
MSE                       & 0.084                  & \textbf{0.153*} & 0.101                                                                & 0.459           & 0.550           & 0.837            & 0.184           & 52.4\%                                                                      & 0.995                                                                                               \\
Binary                    & 0.007                  & 0.155           & 0.103                                                                & 0.779           & 0.045           & 0.068            & 0.009           & 50.2\%                                                                      & 1.001                                                                                               \\ \midrule
{\ul \textbf{LSTM}}       &                        &                 &                                                                      &                 &                 &                  &                 &                                                                             &                                                                                                     \\
Sharpe                    & \textbf{0.451*}        & 0.155           & 0.105                                                                & \textbf{0.209*} & \textbf{2.907*} & \textbf{4.290*}  & \textbf{2.159*} & \textbf{59.6\%*}                                                            & \textbf{1.113*}                                                                                     \\
Ave. Returns              & 0.208                  & 0.154           & 0.102                                                                & 0.365           & 1.349           & 2.045            & 0.568           & 54.8\%                                                                      & 1.028                                                                                               \\
MSE                       & 0.121                  & 0.154           & 0.100                                                                & 0.362           & 0.791           & 1.211            & 0.335           & 52.8\%                                                                      & 1.020                                                                                               \\
Binary                    & 0.075                  & 0.155           & \textbf{0.099*}                                                      & 0.682           & 0.486           & 0.762            & 0.110           & 51.0\%                                                                      & 1.043                                                                                               \\ \bottomrule
\end{tabular}
\end{table*}

From Exhibit \ref{table:performancerescaled}, while the addition of volatility scaling at the portfolio level improved performance ratios on the whole, it had a larger beneficial effect on machine learning models compared to the reference benchmarks -- propelling Sharpe-optimised MLPs to outperform returns-optimised ones, and even leading to Sharpe-optimised linear models beating reference benchmarks. From a risk perspective, we can see that both volatility and downside deviation also become a lot more comparable, with the former hovering close to $15.5\%$ and the latter around $10\%$. However, Sharpe-optimised LSTMs still retained the lowest MDD across all models, with superior risk-adjusted performance ratios across the board. Referring to the cumulative returns plots for the rescaled portfolios in Exhibit \ref{fig:cumulative_returns}, the benefits of direct outputs with Sharpe ratio optimisation can also be observed -- with larger cumulative returns observed for linear, MLP and LSTM models compared to the reference benchmarks. Furthermore, we note the general underperformance of models which use standard regression and classification methods for trend estimation -- hinting at the difficulties faced in selecting an appropriate position sizing function, and in optimising models to generate positions without accounting for risk. This is particularly relevant for binary classification methods, which produce relatively flat equity lines and underperform reference benchmarks in general. Some of these poor results can be explained by the implicit decision threshold adopted. From the percentage of positive returns captured in Exhibit \ref{table:performancerescaled}, most binary classification models have about a $50 \%$ accuracy which, while expected of a classifier with a 0.5 probability threshold, is far below the accuracies seen in other benchmarks. Furthermore, performance is made worse by the fact that the model's magnitude of gains versus losses $\left(\frac{\text{Ave. P}}{\text{Ave. L}}\right)$ is much smaller than competing methods -- with average loss magnitudes even outweighing profits for the MLP classifier $\left(\frac{\text{Ave. P}}{\text{Ave. L}} = 0.986\right)$. As such, these observations lend support to the direct generation of positions sizes with machine learning methods, given the multiple considerations (e.g. decision thresholds and profit/loss magnitudes) that would be required to incorporate standard supervising learning methods into a profitable trading strategy.

\begin{figure*}[bpth]
\caption{Performance Across Individual Assets}
\label{fig:CrossSectionPerformance}
\centering
\begin{subfigure}[]{\linewidth}
\centering
\includegraphics[width=1\linewidth]{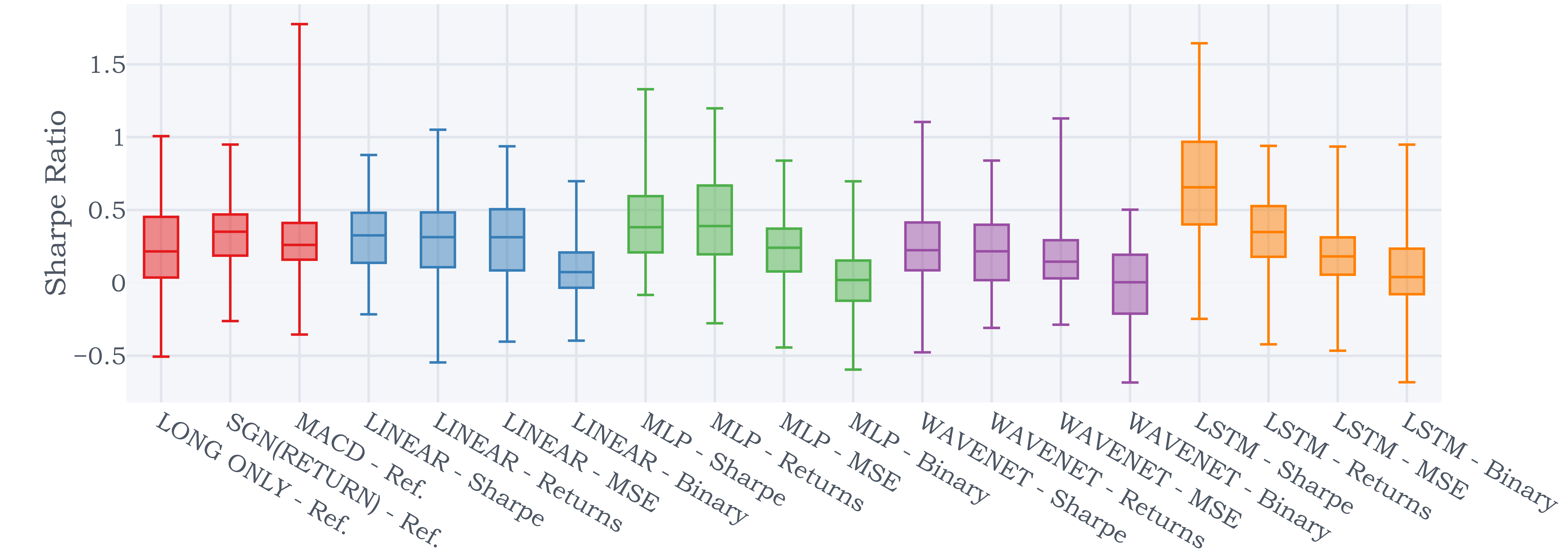}
\caption{Sharpe Ratio}
\label{fig:sharpe_box_plot}
\end{subfigure}\vfill
\begin{subfigure}[]{\linewidth}
\centering
\includegraphics[width=1\linewidth]{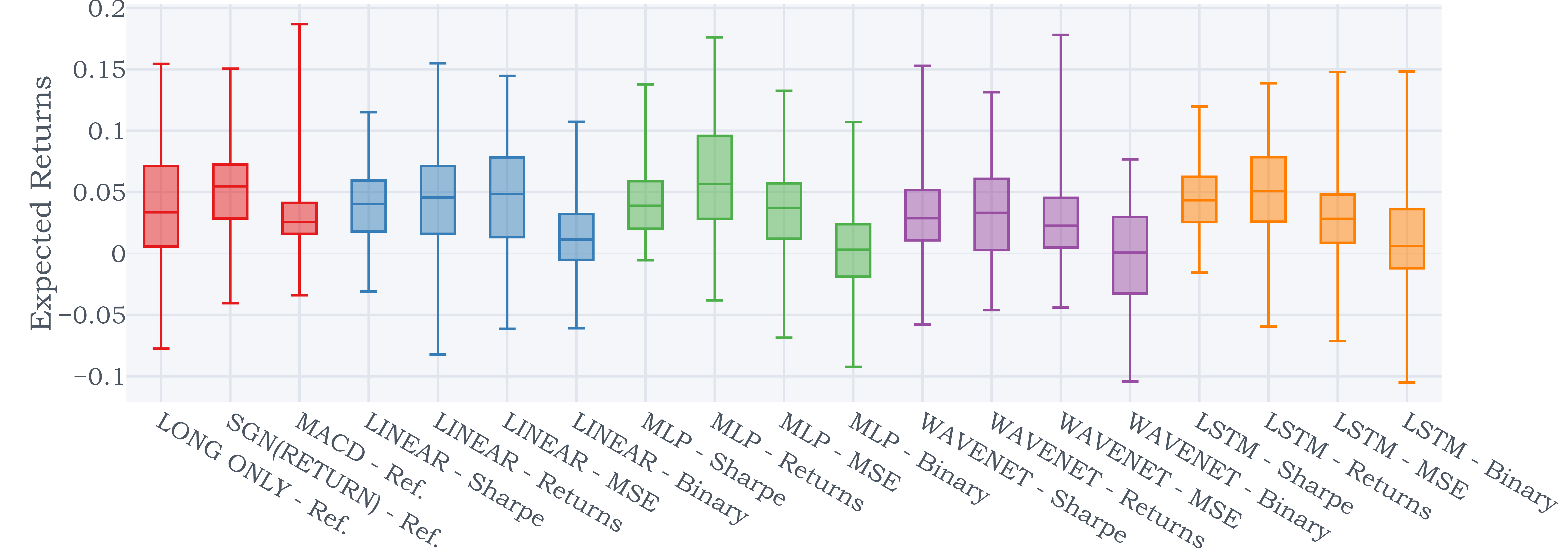}
\caption{Average Returns}
\label{fig:returns_box_plot}
\end{subfigure}\vfill
\begin{subfigure}[]{1\linewidth}
\centering
\includegraphics[width=1\linewidth]{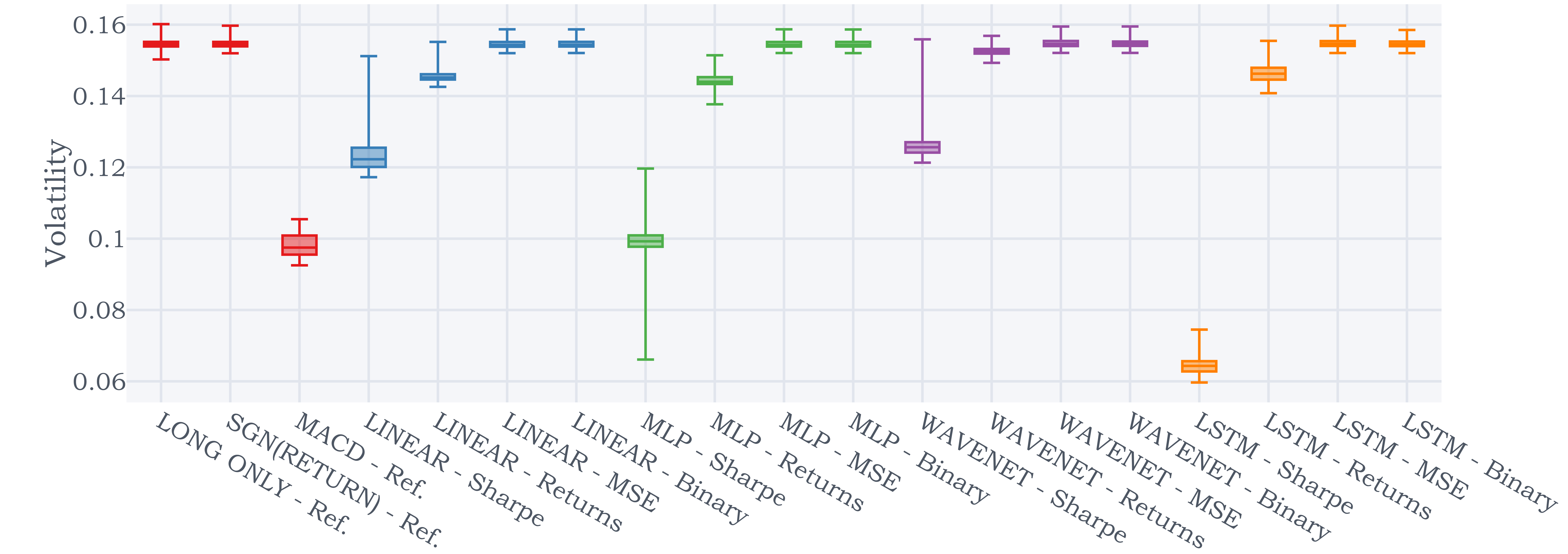}
\caption{Volatility}
\label{fig:vol_box_plot}
\end{subfigure}
\end{figure*}

Strategy performance could also be aided by diversification across a range of assets, particularly when the correlation between signals is low. Hence, to evaluate the raw quality of the underlying signal, we investigate the performance constituents of the time series momentum portfolios  -- using box plots for a variety of performance metrics, plotting the minimum, lower quartile, median, upper quartile, and maximum values across individual futures contracts. We present in Exhibit \ref{fig:CrossSectionPerformance} plots of one metric per category in Section \ref{sec:backtest}, although similar results can be seen for other performance ratios are documented in Appendix \ref{apdx:extra_results}. In general, the Sharpe ratio plots in Exhibit \ref{fig:sharpe_box_plot} echo previous findings, with direct output methods performing better than indirect trend estimation models. However, as seen in Exhibit \ref{fig:vol_box_plot}, this is mainly attributable to significant reduction in signal volatility for the Sharpe-optimised methods, despite a comparable range of average returns in Exhibit \ref{fig:returns_box_plot}. The benefits of retaining the volatility scaling can also be observed, with individual signal volatility capped near the target across all methods -- even with a naive $\sgn(.)$ position sizer. As such, the combination of volatility scaling, direct outputs and Sharpe ratio optimisation were all key to performance gains in Deep Momentum Networks.

\section{Turnover Analysis}
\label{sec:costs}
To investigate how transaction costs affect strategy performance, we first analyse the daily position changes of the signal  --  characterised for asset $i$ by daily turnover $\zeta_t^{(i)}$ as defined in \cite{DemystifyingTimeSeriesMomentum}:
\begin{align}
\zeta_t^{(i)} = \sigma_{\mathrm{tgt}} \left| \frac{X_t^{(i)}}{\sigma_t^{(i)}} -  \frac{X_{t-1}^{(i)}}{\sigma_{t-1}^{(i)}} \right|
\end{align}

Which is broadly proportional to the volume of asset $i$ traded on day $t$ with reference to the updated portfolio weights. 

\begin{figure*}[phb]
\centering
\caption{Turnover Analysis}
\begin{subfigure}[]{\linewidth}
\centering
\includegraphics[width=0.9\linewidth]{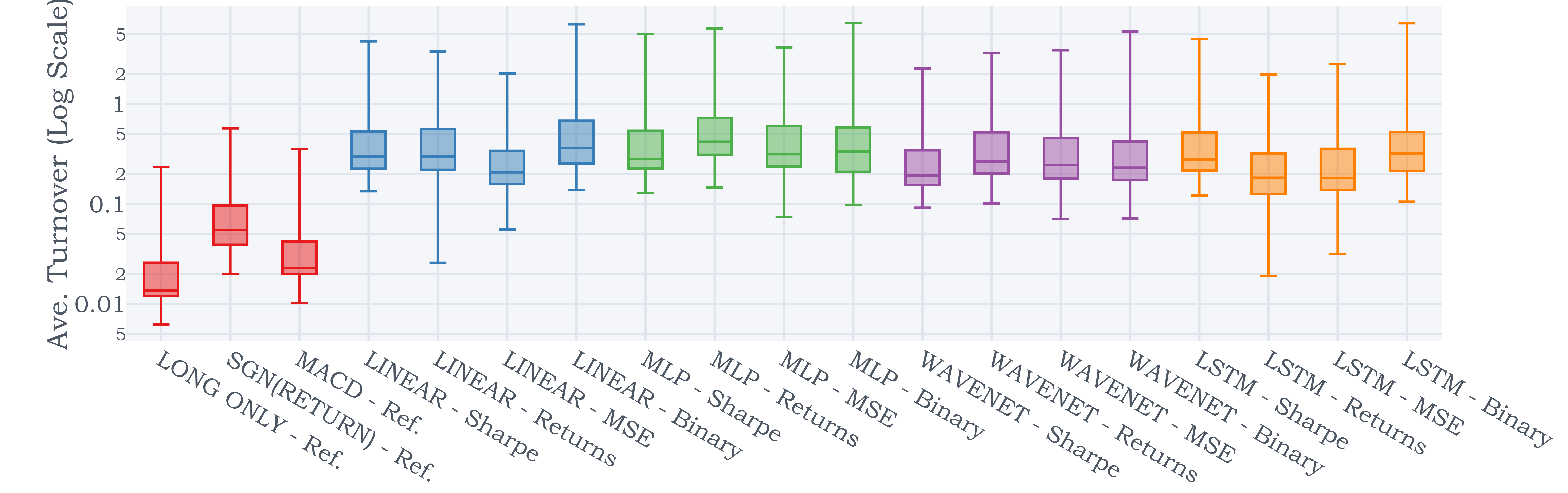}
\caption{Average Strategy Turnover}
\label{fig:turnover}
\end{subfigure}
\begin{subfigure}[]{\linewidth}
\centering
\includegraphics[width=0.9\linewidth]{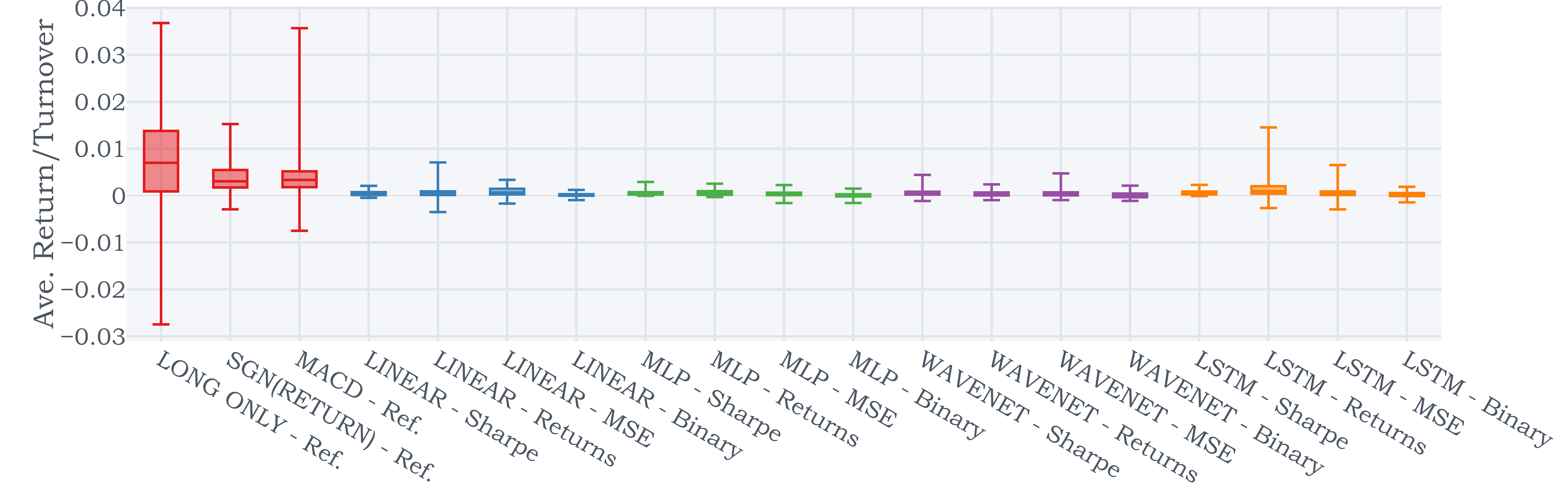}
\caption{Average Returns / Average Turnover}
\label{fig:rets_v_turnover}
\end{subfigure}
\end{figure*}

Exhibit \ref{fig:turnover} shows the average strategy turnover across all assets from 1995 to 2015, focusing on positions generated by the raw signal outputs. As the box plots are charted on a logarithm scale, we note that while the machine learning-based models have a similar turnover, they also trade significantly more than the reference benchmarks -- approximately 10 times more compared to the Long Only benchmark. This is also reflected in Exhibit \ref{fig:turnover} which compares the average daily returns against the average daily turnover -- with ratios from machine learning models lying close to the x-axis. 

\begin{figure*}[phtb]
\caption{Impact of Transaction Costs on Sharpe Ratio}
\label{fig:ex_cost_sharpe}
\centering
\includegraphics[width=0.925\linewidth]{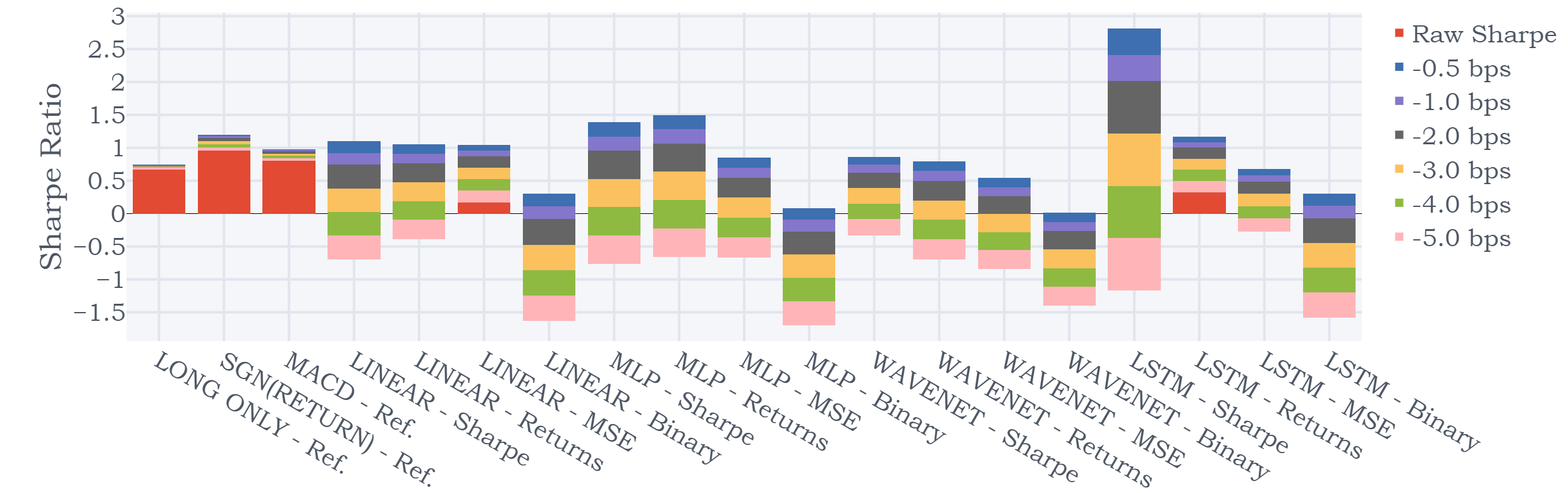}
\end{figure*}

To concretely quantify the impact of transaction costs on performance, we also compute the ex-cost Sharpe ratios -- using the rebalancing costs defined in \cite{DemystifyingTimeSeriesMomentum} to adjust our returns for a variety of transaction cost assumptions . For the results in Exhibit \ref{fig:ex_cost_sharpe}, the top of each bar chart marks the maximum cost-free Sharpe ratio of the strategy, with each coloured block denoting the Sharpe ratio reduction for the corresponding cost assumption. In line with the turnover analysis, the reference benchmarks demonstrate the most resilience to high transaction costs (up to 5bps), with the profitability across most machine learning models persisting only up to 4bps. However, we still obtain higher cost-adjusted Sharpe ratios with the Sharpe-optimised LSTM for up to 2-3 bps, demonstrating its suitability for trading more liquid instruments. 


\begin{table*}[phbt]
\centering
\caption{Performance Metrics with Transaction Costs ($c=10$bps)}
\label{tab:performance_with_costs}
\begin{tabular}{@{}llllllllll@{}}
\toprule
\textbf{}     & \textbf{E[Return]} & \textbf{Vol.} & \textbf{\begin{tabular}[c]{@{}l@{}}Downside\\     Deviation\end{tabular}} & \textbf{MDD} & \textbf{Sharpe} & \textbf{Sortino} & \textbf{Calmar} & \textbf{\begin{tabular}[c]{@{}l@{}}\% of $+$ve \\     Returns\end{tabular}} & \textbf{\begin{tabular}[c]{@{}l@{}}$\mathbf{\frac{\text{Ave. P}}{\text{Ave. L}}}$\end{tabular}} \\ \hline
Long Only     & 0.097              & \textbf{0.154*}         & 0.103                                                                     & 0.482        & 0.628           & 0.942            & 0.201           & 53.3\%                                                                       & 0.970                                                                                               \\
Sgn(Returns) & 0.133              & \textbf{0.154*}         & \textbf{0.102*}                                                                     & 0.373        & 0.861           & 1.296            & 0.356           & 53.3\%                                                                       & 1.011                                                                                               \\
MACD          & 0.111              & 0.155         & 0.106                                                                     & 0.472        & 0.719           & 1.047            & 0.236           & 52.5\%                                                                       & \textbf{1.020*}                                                                                               \\ \midrule
LSTM          & -0.833             & 0.157         & 0.114                                                                     & 1.000        & -5.313          & -7.310           & -0.833          & 33.9\%                                                                       & 0.793                                                                                               \\
LSTM + Reg.   & \textbf{0.141*}              & \textbf{0.154*}         & \textbf{0.102*}                                                                    & \textbf{0.371*}        & \textbf{0.912*}          & \textbf{1.379*}            & \textbf{0.379*}           & \textbf{53.4\%*}                                                                      & 1.014                                                                                               \\ \hline
\end{tabular}
\end{table*}

\subsection{Turnover Regularisation}
One simple way to account for transaction costs is to use cost-adjusted returns $\tilde{r}_{t,t+1}^{TSMOM}$ directly during training, augmenting the strategy returns defined in Equation \eqref{eqn:tsmom} as below:

\begin{align}
&\tilde{r}_{t,t+1}^{TSMOM} = \nonumber \\ 
&\frac{\sigma_{\mathrm{tgt}}}{N_t} \sum_{i=1}^{N_t} \bigg( \frac{X_t^{(i)}}{\sigma_t^{(i)}}~r_{t,t+1}^{(i)} -  c \left| \frac{X_t^{(i)}}{\sigma_t^{(i)}} -  \frac{X_{t-1}^{(i)}}{\sigma_{t-1}^{(i)}} \right| \bigg),
\end{align}
where $c$ is a constant reflecting transaction cost assumptions. As such, using $\tilde{r}_{t,t+1}^{TSMOM}$ in Sharpe ratio loss functions during training corresponds to optimising the ex-cost risk-adjusted returns, and $c \left| \frac{X_t^{(i)}}{\sigma_t^{(i)}} -  \frac{X_{t-1}^{(i)}}{\sigma_{t-1}^{(i)}} \right|$ can also be interpreted as a regularisation term for turnover.

Given that the Sharpe-optimised LSTM is still profitable in the presence of small transactions costs, we seek to quantify the effectiveness of turnover regularisation when costs are prohibitively high -- considering the extreme case where $c=10$bps in our investigation. Tests were focused on the Sharpe-optimised LSTM with and without the turnover regulariser (LSTM + Reg. for the former) -- including the additional portfolio level volatility scaling to bring signal volatilities to the same level. Based on the results in Exhibit \ref{tab:performance_with_costs}, we can see that the turnover regularisation does help improve the LSTM in the presence of large costs, leading to slightly better performance ratios when compared to the reference benchmarks.

\section{Conclusions}
We introduce Deep Momentum Networks -- a hybrid class of deep learning models which retain the volatility scaling framework of time series momentum strategies while using deep neural networks to output  position targeting trading signals. Two approaches to position generation were evaluated here. Firstly, we cast trend estimation as a standard supervised learning problem -- using machine learning models to forecast the expected asset returns or probability of a positive return at the next time step -- and apply a simple maximum long/short trading rule based on the direction of the next return. Secondly, trading rules were directly generated as outputs from the model, which we calibrate by maximising the Sharpe ratio or average strategy return. Testing this on a universe of continuous futures contracts, we demonstrate clear improvements in risk-adjusted performance by calibrating models with the Sharpe ratio -- where the LSTM model achieved best results. Incorporating transaction costs, the Sharpe-optimised LSTM outperforms benchmarks up to 2-3 basis points of costs, demonstrating its suitability for trading more liquid assets. To accommodate high costs settings, we introduce a turnover regulariser to use during training, which was shown to be effective even in extreme scenarios (i.e. $c=10$bps).

Future work includes extensions of the framework presented here to incorporate ways to deal better with non-stationarity in the data, such as using the recently introduced Recurrent Neural Filters \cite{rnf}. Another direction of future work focuses on the study of time series momentum at the microstructure level.

\section{Acknowledgements}
We would like to thank Anthony Ledford, James Powrie and Thomas Flury for their interesting comments as well the Oxford-Man Institute of Quantitative Finance for financial support.

% References Section.
\newpage
\bibliographystyle{IEEEtran}
{\footnotesize
\bibliography{mom_bib}
} 

\newpage
\appendix
\subsection{Dataset Details}
\label{apdx:data}
From the full 98 ratio-adjusted continuous futures contracts in the Pinnacle Data Corp CLC Database, we extract 88 which have $<10\%$ of its data missing -- with a breakdown by asset class below:

\subsubsection{Commodities}~
% \usepackage{booktabs}
\begin{table}[h]
\centering
\begin{tabular}{ll}
\midrule
\textbf{Identifier} & \textbf{Description}       \\ \midrule
BC                  & BRENT CRUDE OIL, composite \\
BG                  & BRENT GASOIL, comp.        \\
BO                  & SOYBEAN OIL                \\
CC                  & COCOA                      \\
CL                  & CRUDE OIL                  \\
CT                  & COTTON \#2                 \\
C\_                 & CORN                       \\
DA                  & MILK III, Comp.            \\
FC                  & FEEDER CATTLE              \\
GC                  & GOLD (COMMEX)              \\
GI                  & GOLDMAN SAKS C. I.         \\
HG                  & COPPER                     \\ 
HO                  & HEATING OIL \#2            \\
JO                  & ORANGE JUICE               \\
KC                  & COFFEE                     \\
KW                  & WHEAT, KC                  \\
LB                  & LUMBER                     \\
LC                  & LIVE CATTLE                \\
LH                  & LIVE HOGS                  \\
MW                  & WHEAT, MINN                \\
NG                  & NATURAL GAS                \\
NR                  & ROUGH RICE                 \\
O\_                 & OATS                       \\
PA                  & PALLADIUM                  \\
PL                  & PLATINUM                   \\
RB                  & RBOB GASOLINE              \\
SB                  & SUGAR \#11                 \\
SI                  & SILVER  (COMMEX)           \\
SM                  & SOYBEAN MEAL               \\
S\_                 & SOYBEANS                   \\
W\_                 & WHEAT, CBOT                \\
ZA                  & PALLADIUM, electronic      \\
ZB                  & RBOB, Electronic           \\
ZC                  & CORN, Electronic           \\
ZF                  & FEEDER CATTLE, Electronic  \\
ZG                  & GOLD, Electronic           \\
ZH                  & HEATING OIL, electronic    \\
ZI                  & SILVER, Electronic         \\
ZK                  & COPPER, electronic         \\
ZL                  & SOYBEAN OIL, Electronic    \\
ZM                  & SOYBEAN MEAL, Electronic   \\
ZN                  & NATURAL GAS, electronic    \\
ZO                  & OATS, Electronic           \\
ZP                  & PLATINUM, electronic       \\
ZR                  & ROUGH RICE, Electronic     \\
ZS                  & SOYBEANS, Electronic       \\
ZT                  & LIVE CATTLE, Electronic    \\
ZU                  & CRUDE OIL, Electronic      \\
ZW                  & WHEAT, Electronic          \\
ZZ                  & LEAN HOGS, Electronic     \\ \hline
\end{tabular}
\end{table}

\subsubsection{Equities}~

% \usepackage{booktabs}
\begin{table}[h]
\centering
\begin{tabular}{@{}ll@{}}
\toprule
\textbf{Identifier} & \textbf{Description}       \\ \midrule
AX                  & GERMAN DAX INDEX           \\
CA                  & CAC40 INDEX                \\
EN                  & NASDAQ, MINI               \\
ER                  & RUSSELL 2000, MINI         \\
ES                  & S \& P 500, MINI           \\
HS                  & HANG SENG                  \\
LX                  & FTSE 100 INDEX             \\
MD                  & S\&P 400 (Mini electronic) \\
SC                  & S \& P 500, composite      \\
SP                  & S \& P 500, day session    \\
XU                  & DOW JONES EUROSTOXX50      \\
XX                  & DOW JONES STOXX 50         \\
YM                  & Mini Dow Jones (\$5.00)    \\ \bottomrule
\end{tabular}
\end{table}

\subsubsection{Fixed Income}~
% \usepackage{booktabs}
\begin{table}[h]
\centering
\begin{tabular}{@{}ll@{}}
\toprule
\textbf{Identifier} & \textbf{Description}     \\ \midrule
AP                  & AUSTRALIAN PRICE INDEX   \\
DT                  & EURO BOND (BUND)         \\
FA                  & T-NOTE, 5yr day session  \\
FB                  & T-NOTE, 5yr composite    \\
GS                  & GILT, LONG  BOND         \\
TA                  & T-NOTE, 10yr day session \\
TD                  & T-NOTES, 2yr day session \\
TU                  & T-NOTES, 2yr composite   \\
TY                  & T-NOTE, 10yr composite   \\
UA                  & T-BONDS, day session     \\
UB                  & EURO BOBL                \\
US                  & T-BONDS, composite       \\ \bottomrule
\end{tabular}
\end{table}

\subsubsection{FX}~

% \usepackage{booktabs}
\begin{table}[h]
\centering
\begin{tabular}{@{}ll@{}} \toprule
\textbf{Identifier} & \textbf{Description} \\ \midrule
AD & AUSTRALIAN \$\$, day session \\
AN & AUSTRALIAN \$\$, composite   \\
BN & BRITISH POUND, composite   \\
CB & CANADIAN 10YR BOND         \\
CN & CANADIAN \$\$, composite     \\
DX & US DOLLAR INDEX            \\
FN & EURO, composite            \\
FX & EURO, day session          \\
JN & JAPANESE YEN, composite    \\
MP & MEXICAN PESO               \\
NK & NIKKEI INDEX               \\
SF & SWISS FRANC, day session   \\
SN & SWISS FRANC, composite     \\ \bottomrule
\end{tabular}
\end{table}

To reduce the impact of outliers, we also winsorise the data by capping/flooring it to be within 5 times its exponentially weighted moving (EWM) standard deviations from its EWM average -- computed using a 252-day half life.
\subsection{Hyperparameter Optimisation}
\label{apdx:hyperparam}
% \usepackage{booktabs}
\begin{table*}[]
\centering
\caption{Hyperparameter Search Range}
\label{tab:hyperparams}
\begin{tabular}{@{}lll@{}}
\toprule
\textbf{Hyperparameters}           & \textbf{Random Search Grid}                                & \textbf{Notes}        \\ \midrule
Dropout Rate                       & 0.1, 0.2, 0.3, 0.4, 0.5                                    & Neural Networks Only  \\
Hidden Layer Size                  & 5, 10, 20, 40, 80                                          & Neural Networks Only  \\
Minibatch Size                     & 256, 512, 1024, 2048                                       &                       \\
Learning Rate                      & $10^{-5},~ 10^{-4},~ 10^{-3},~ 10^{-2},~ 10^{-1},~ 10^{0}$ &                       \\
Max Gradient Norm                  & $10^{-4},~ 10^{-3},~ 10^{-2},~ 10^{-1},~ 10^{0},~ 10^{1}$  &                       \\
L1 Regularisation Weight ($\alpha$) & $10^{-5},~ 10^{-4},~ 10^{-3},~ 10^{-2},~ 10^{-1}$          & Lasso Regression Only \\ \bottomrule
\end{tabular}
\end{table*}

Hyperparameter optimisation was applied using 50 iterations of random search, with the full search grid documented in Exhibit \ref{tab:hyperparams}, with the models fully recalibrated every 5 years using all available data up to that point. For LSTM-based models, time series were subdivided into trajectories of 63 time steps ($\approx$ 3 months), with the LSTM unrolled across the length of the trajectory during backpropagation.

\subsection{Additional Results}
\label{apdx:extra_results}
In addition to the selected results in Section \ref{sec:tests}, we also present a full list of results was presented for completeness -- echoing the key findings reported in the discuss. Detailed descriptions of the plots and tables can be found below:

\subsubsection{Cross-Validation Performance} The testing procedure in Section \ref{sec:tests} can also be interpreted as a cross-validation approach -- splitting the original dataset into six 5-year blocks (1990-2015), calibrating using an expanding window of data, and testing out-of-sample on the next block outside the training set. As such, for consistency with machine learning literature, we present our results in a cross validation format as well -- reporting the average value across all blocks $\pm$ 2 standard deviations. Furthermore, this also gives an indication of how signal performance varies across the various time periods.

\begin{itemize}
\item \textbf{Exhibit \ref{table:crossvalidraw}} -- Cross-validation results for raw signal outputs.
\item \textbf{Exhibit \ref{table:crossvalidrescaled}} -- Cross-validation results for signals which have been rescaled to target volatility at the portfolio level.
\end{itemize}

\subsubsection{Metrics Across Individual Assets} 
\begin{figure*}[bpth]
\caption{Performance Ratios Across Individual Assets}
\label{fig:apdx_asset_level_perf_ratios}
\centering
\begin{subfigure}[]{\linewidth}
\includegraphics[width=1\linewidth]{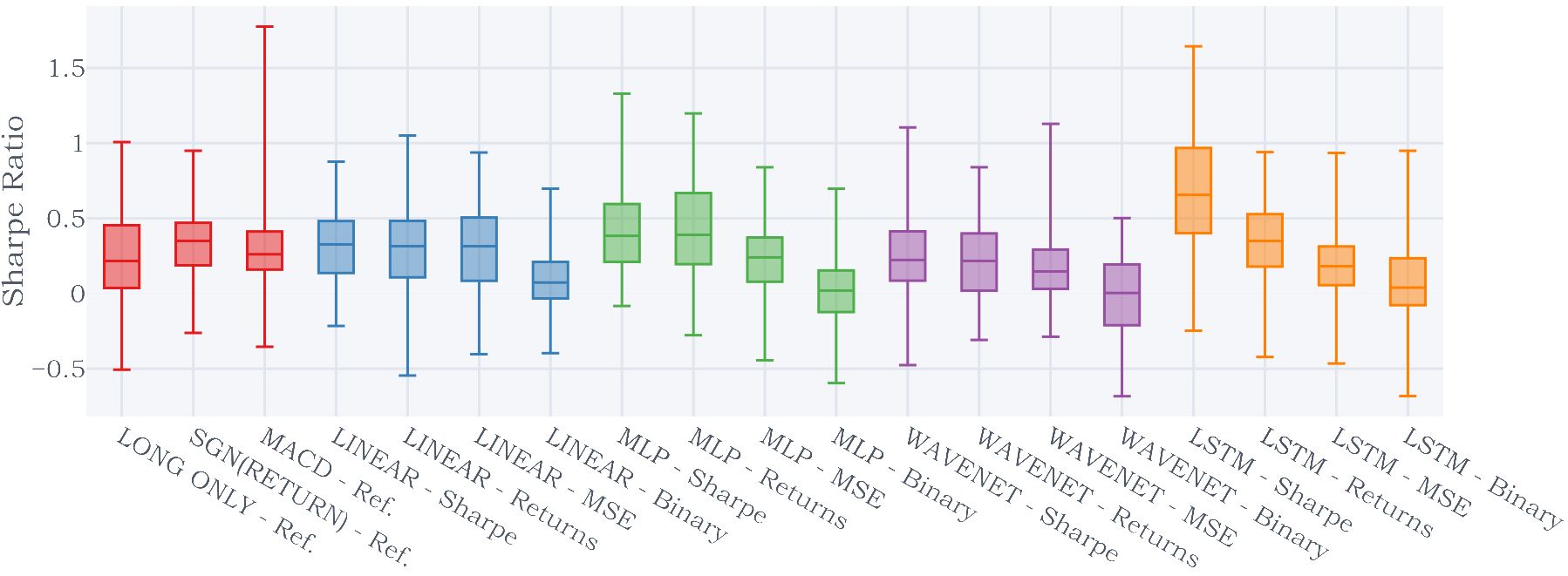}
\caption{Sharpe Ratio}
\end{subfigure}\\
\begin{subfigure}[]{1\linewidth}
\includegraphics[width=1\linewidth]{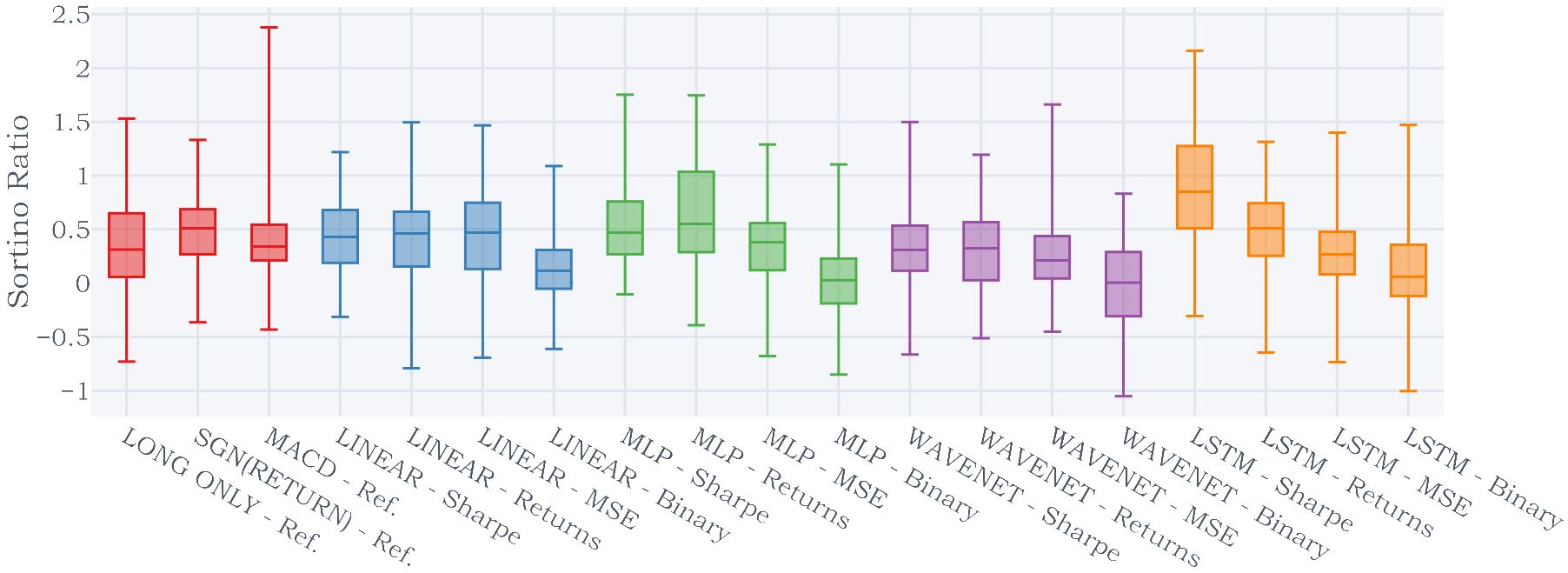}
\caption{Sortino Ratio}
\end{subfigure}
\begin{subfigure}[]{1\linewidth}
\includegraphics[width=1\linewidth]{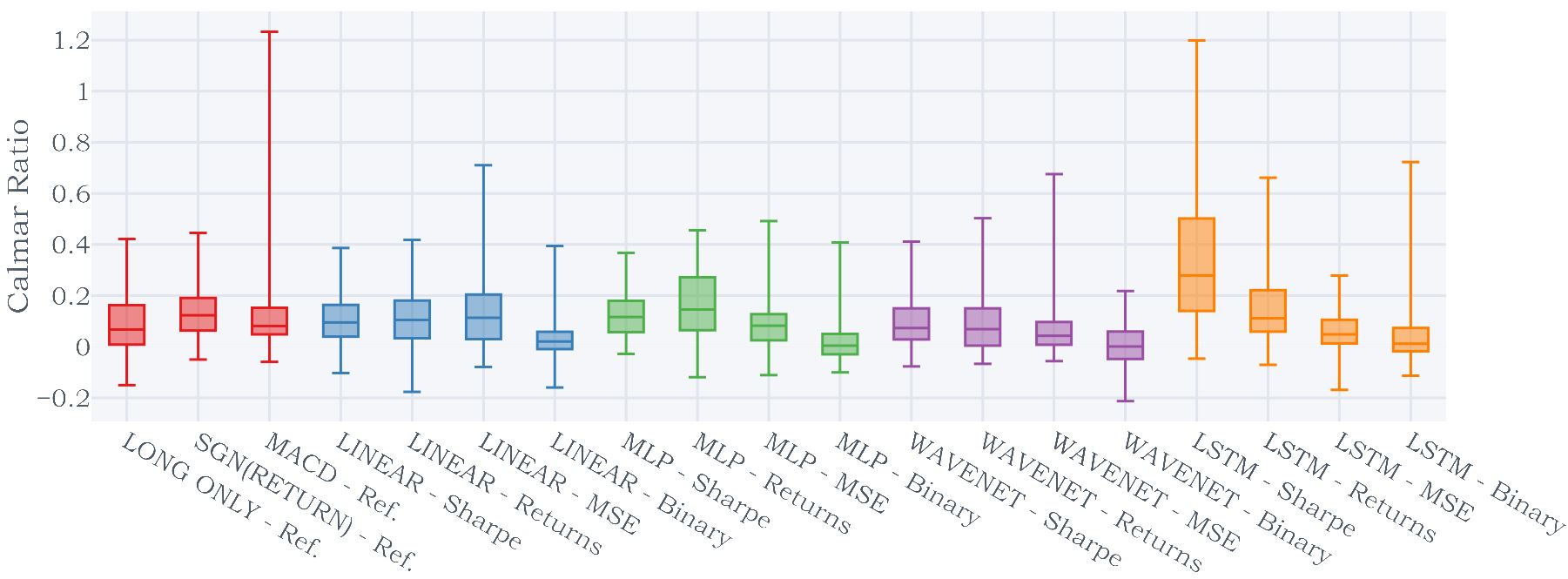}
\caption{Calmar Ratio}
\end{subfigure} 
\end{figure*}

\begin{figure*}[bpth]
\caption{Reward vs Risk Across Individual Assets}
\label{fig:apdx_asset_level_risk_v_reward}
\centering
\begin{subfigure}[]{\linewidth}
\includegraphics[width=1\linewidth]{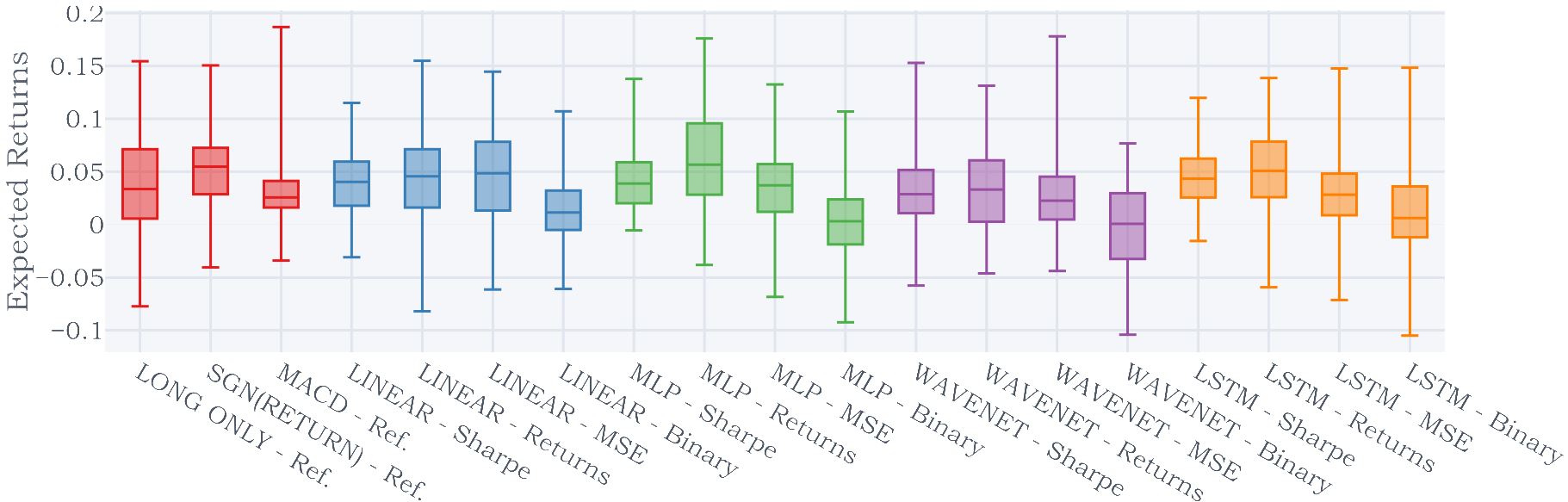}
\caption{Expected Returns}
\end{subfigure}
\begin{subfigure}[]{1\linewidth}
\includegraphics[width=1\linewidth]{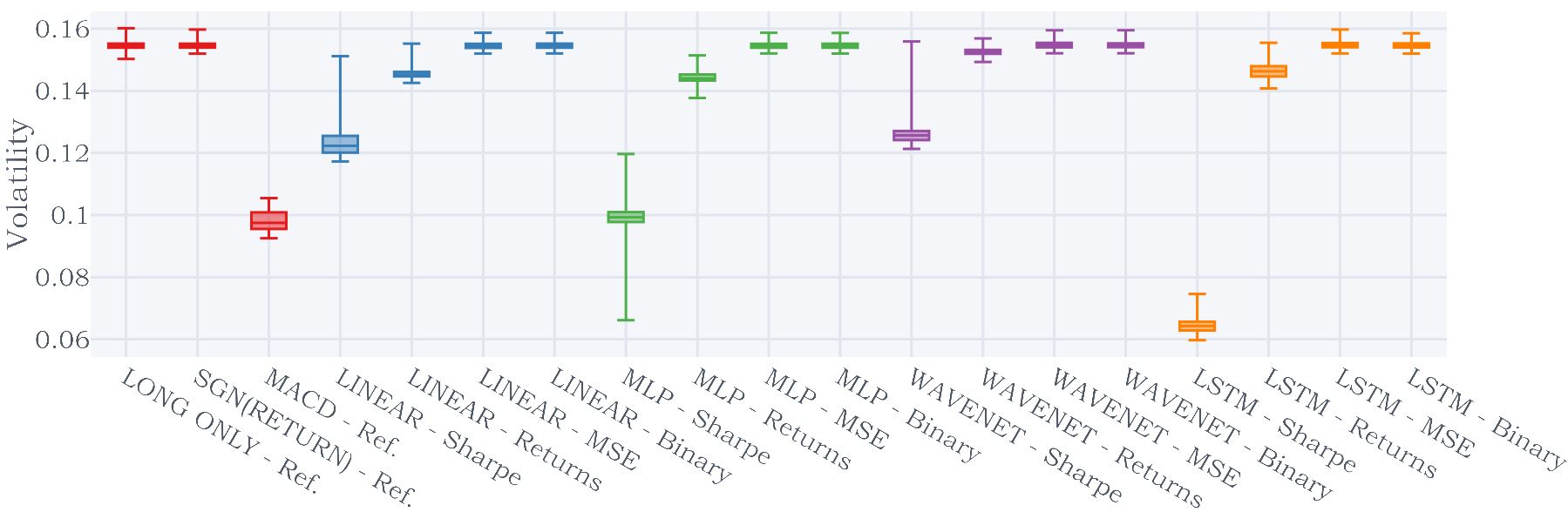}
\caption{Volatility}
\end{subfigure}
\begin{subfigure}[]{1\linewidth}
\includegraphics[width=1\linewidth]{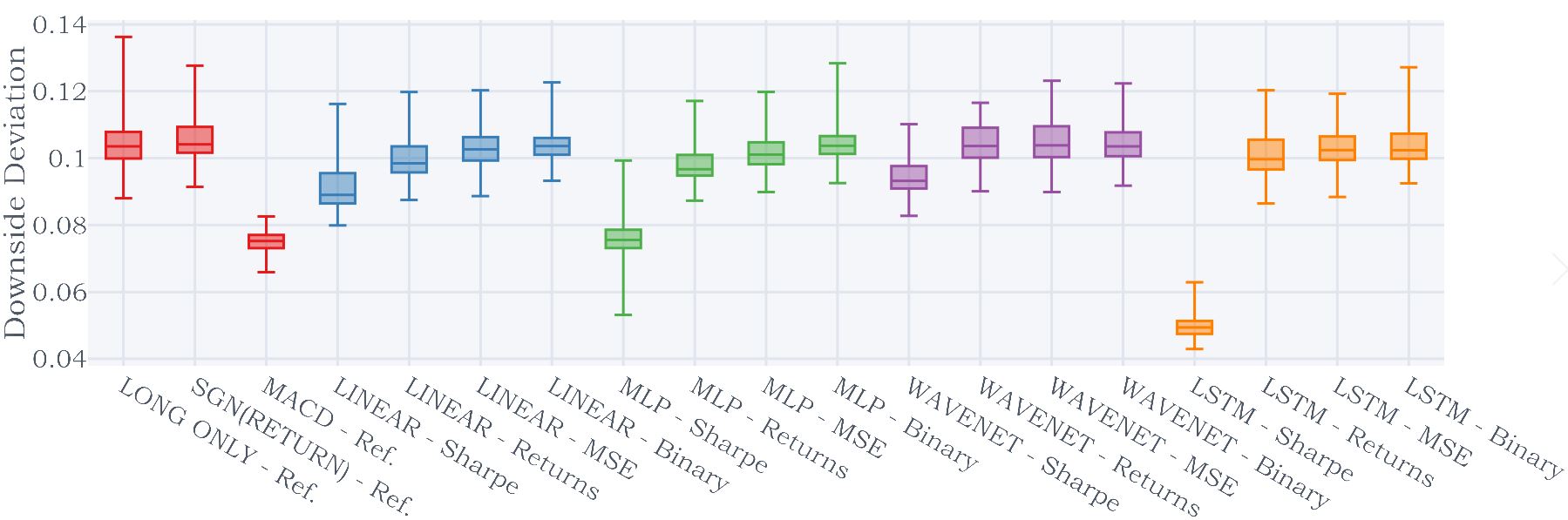}
\caption{Downside Deviation}
\end{subfigure} 
\begin{subfigure}[]{1\linewidth}
\includegraphics[width=1\linewidth]{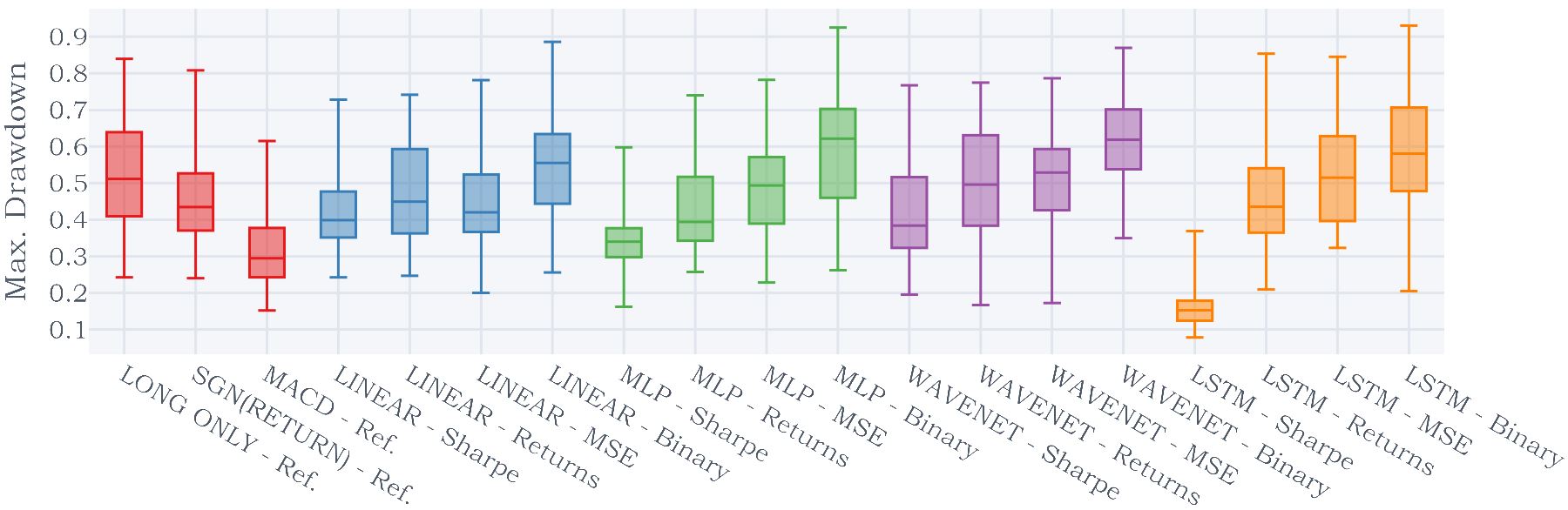}
\caption{Max. Drawdown}
\end{subfigure} 
\end{figure*}

~ We also provide additional plots on performance of other risk metrics and performance ratios across individual assets, as described below:

\begin{itemize}
\item \textbf{Exhibit \ref{fig:apdx_asset_level_perf_ratios}} -- Full performance ratios (Sharpe, Sortino, Calmar).
\item \textbf{Exhibit \ref{fig:apdx_asset_level_risk_v_reward}} -- Box plots including additional risk metrics (volatility, downside deviation and maximum drawdown). We note that our findings for other risk metrics are similar to volatility -- with Sharpe-optimised models lowering risk across different methods.
\end{itemize}

\begin{table*}[htbp]
\centering
\caption{Cross-Validation Performance -- Raw Signal Outputs}
\label{table:crossvalidraw}
\begin{tabular}{@{}lllll@{}}
\toprule
\textbf{}                 & \textbf{E[Return]}           & \textbf{Vol.}                & \textbf{\begin{tabular}[c]{@{}l@{}}Downside\\     Deviation\end{tabular}} & \textbf{MDD}                 \\ \midrule
{\ul \textbf{Reference}} &                              &                              &                                                                           &                              \\
Long Only                 & 0.043 $\pm$ 0.028            & 0.054 $\pm$ 0.016            & 0.037 $\pm$ 0.013                                                         & 0.116 $\pm$ 0.091            \\
Sgn(Returns)             & 0.047 $\pm$ 0.051            & 0.046 $\pm$ 0.012            & 0.032 $\pm$ 0.007                                                         & 0.067 $\pm$ 0.041            \\
MACD                      & 0.026 $\pm$ 0.032            & 0.032 $\pm$ 0.008            & 0.023 $\pm$ 0.007                                                         & 0.054 $\pm$ 0.048            \\ \midrule
{\ul \textbf{Linear}}     &                              &                              &                                                                           &                              \\
Sharpe                    & 0.034 $\pm$ 0.030            & 0.039 $\pm$ 0.028            & 0.028 $\pm$ 0.020                                                         & 0.072 $\pm$ 0.096            \\
Ave. Returns              & 0.033 $\pm$ 0.031            & 0.046 $\pm$ 0.025            & 0.031 $\pm$ 0.018                                                         & 0.110 $\pm$ 0.114            \\
MSE                       & 0.047 $\pm$ 0.038            & 0.049 $\pm$ 0.019            & 0.033 $\pm$ 0.013                                                         & 0.100 $\pm$ 0.121            \\
Binary                    & 0.012 $\pm$ 0.028            & 0.045 $\pm$ 0.011            & 0.031 $\pm$ 0.009                                                         & 0.109 $\pm$ 0.045            \\ \midrule
{\ul \textbf{MLP}}        &                              &                              &                                                                           &                              \\
Sharpe                    & 0.038 $\pm$ 0.027            & 0.030 $\pm$ 0.041            & 0.021 $\pm$ 0.028                                                         & 0.062 $\pm$ 0.160            \\
Ave. Returns              & \textbf{0.056  $\pm$ 0.046*} & 0.044 $\pm$ 0.024            & 0.030 $\pm$ 0.017                                                         & 0.075 $\pm$ 0.150            \\
MSE                       & 0.037 $\pm$ 0.051            & 0.048 $\pm$ 0.021            & 0.032 $\pm$ 0.015                                                         & 0.109 $\pm$ 0.134            \\
Binary                    & -0.004 $\pm$ 0.028           & 0.042 $\pm$ 0.007            & 0.028 $\pm$ 0.006                                                         & 0.111 $\pm$ 0.079            \\ \midrule
{\ul \textbf{WaveNet}}    &                              &                              &                                                                           &                              \\
Sharpe                    & 0.030 $\pm$ 0.030            & 0.038 $\pm$ 0.019            & 0.027 $\pm$ 0.015                                                         & 0.069 $\pm$ 0.055            \\
Ave. Returns              & 0.034 $\pm$ 0.043            & 0.042 $\pm$ 0.003            & 0.030 $\pm$ 0.003                                                         & 0.088 $\pm$ 0.062            \\
MSE                       & 0.024 $\pm$ 0.046            & 0.043 $\pm$ 0.010            & 0.030 $\pm$ 0.010                                                         & 0.102 $\pm$ 0.056            \\
Binary                    & -0.009 $\pm$ 0.023           & 0.043 $\pm$ 0.008            & 0.030 $\pm$ 0.008                                                         & 0.159 $\pm$ 0.107            \\ \midrule
{\ul \textbf{LSTM}}       &                              &                              &                                                                           &                              \\
Sharpe                    & 0.045 $\pm$ 0.030            & \textbf{0.017  $\pm$ 0.004*} & \textbf{0.012  $\pm$ 0.003*}                                              & \textbf{0.019  $\pm$ 0.005*} \\
Ave. Returns              & 0.045 $\pm$ 0.050            & 0.048 $\pm$ 0.018            & 0.034 $\pm$ 0.011                                                         & 0.104 $\pm$ 0.119            \\
MSE                       & 0.023 $\pm$ 0.037            & 0.048 $\pm$ 0.022            & 0.033 $\pm$ 0.017                                                         & 0.116 $\pm$ 0.082            \\
Binary                    & -0.005 $\pm$ 0.088           & 0.042 $\pm$ 0.003            & 0.027 $\pm$ 0.006                                                         & 0.151 $\pm$ 0.211            \\ \bottomrule
\end{tabular}
\begin{tabular}{@{}llllll@{}}
\\\\
\toprule
\textbf{}                 & \textbf{Sharpe}              & \textbf{Sortino}             & \textbf{Calmar}              & \textbf{\begin{tabular}[c]{@{}l@{}}Fraction of \\ $+$ve   Returns\end{tabular}} & \textbf{\begin{tabular}[c]{@{}l@{}}$\mathbf{\frac{\text{Ave.  P}}{\text{Ave. L}}}$\end{tabular}} \\ \midrule
{\ul \textbf{Reference}} &                              &                              &                              &                                                                             &                                                                                                     \\
Long Only                 & 0.839 $\pm$ 0.786            & 1.258 $\pm$ 1.262            & 0.420 $\pm$ 0.490            & 0.546 $\pm$ 0.025                                                           & 0.956 $\pm$ 0.135                                                                                   \\
Sgn(Returns)             & 1.045 $\pm$ 1.230            & 1.528 $\pm$ 1.966            & 0.864 $\pm$ 1.539            & 0.543 $\pm$ 0.061                                                           & 1.002 $\pm$ 0.067                                                                                   \\
MACD                      & 0.839 $\pm$ 1.208            & 1.208 $\pm$ 1.817            & 0.625 $\pm$ 1.033            & 0.532 $\pm$ 0.030                                                           & 1.016 $\pm$ 0.079                                                                                   \\ \midrule
{\ul \textbf{Linear}}     &                              &                              &                              &                                                                             &                                                                                                     \\
Sharpe                    & 1.025 $\pm$ 1.530            & 1.451 $\pm$ 2.154            & 0.800 $\pm$ 1.772            & 0.544 $\pm$ 0.042                                                           & 1.000 $\pm$ 0.100                                                                                   \\
Ave. Returns              & 0.757 $\pm$ 0.833            & 1.150 $\pm$ 1.378            & 0.397 $\pm$ 0.686            & 0.530 $\pm$ 0.009                                                           & 1.005 $\pm$ 0.100                                                                                   \\
MSE                       & 1.012 $\pm$ 1.126            & 1.532 $\pm$ 1.811            & 0.708 $\pm$ 1.433            & 0.540 $\pm$ 0.024                                                           & 1.008 $\pm$ 0.096                                                                                   \\
Binary                    & 0.288 $\pm$ 0.729            & 0.434 $\pm$ 1.113            & 0.123 $\pm$ 0.313            & 0.506 $\pm$ 0.027                                                           & 1.024 $\pm$ 0.051                                                                                   \\ \midrule
{\ul \textbf{MLP}}        &                              &                              &                              &                                                                             &                                                                                                     \\
Sharpe                    & 1.669 $\pm$ 2.332            & 2.420 $\pm$ 3.443            & 1.665 $\pm$ 2.738            & 0.554 $\pm$ 0.063                                                           & 1.069 $\pm$ 0.151                                                                                   \\
Ave. Returns              & 1.415 $\pm$ 1.781            & 2.127 $\pm$ 2.996            & 1.520 $\pm$ 2.761            & 0.553 $\pm$ 0.043                                                           & 1.022 $\pm$ 0.134                                                                                   \\
MSE                       & 0.821 $\pm$ 1.334            & 1.270 $\pm$ 2.160            & 0.652 $\pm$ 1.684            & 0.525 $\pm$ 0.025                                                           & 1.036 $\pm$ 0.127                                                                                   \\
Binary                    & -0.099 $\pm$ 0.648           & -0.180 $\pm$ 0.956           & -0.013 $\pm$ 0.304           & 0.500 $\pm$ 0.042                                                           & 0.986 $\pm$ 0.064                                                                                   \\ \midrule
{\ul \textbf{WaveNet}}    &                              &                              &                              &                                                                             &                                                                                                     \\
Sharpe                    & 0.780 $\pm$ 0.538            & 1.118 $\pm$ 0.854            & 0.477 $\pm$ 0.610            & 0.535 $\pm$ 0.022                                                           & 0.990 $\pm$ 0.094                                                                                   \\
Ave. Returns              & 0.809 $\pm$ 1.113            & 1.160 $\pm$ 1.615            & 0.501 $\pm$ 1.036            & 0.543 $\pm$ 0.059                                                           & 0.963 $\pm$ 0.069                                                                                   \\
MSE                       & 0.513 $\pm$ 0.991            & 0.744 $\pm$ 1.477            & 0.276 $\pm$ 0.509            & 0.527 $\pm$ 0.033                                                           & 0.979 $\pm$ 0.077                                                                                   \\
Binary                    & -0.220 $\pm$ 0.523           & -0.329 $\pm$ 0.768           & -0.043 $\pm$ 0.145           & 0.499 $\pm$ 0.011                                                           & 0.969 $\pm$ 0.043                                                                                   \\ \midrule
{\ul \textbf{LSTM}}       &                              &                              &                              &                                                                             &                                                                                                     \\
Sharpe                    & \textbf{2.781  $\pm$ 2.081*} & \textbf{3.978  $\pm$ 3.160*} & \textbf{2.488  $\pm$ 1.921*} & \textbf{0.593  $\pm$ 0.054*}                                                & \textbf{1.104  $\pm$ 0.199*}                                                                        \\
Ave. Returns              & 0.961 $\pm$ 1.268            & 1.397 $\pm$ 1.926            & 0.679 $\pm$ 1.552            & 0.547 $\pm$ 0.039                                                           & 0.972 $\pm$ 0.118                                                                                   \\
MSE                       & 0.451 $\pm$ 0.526            & 0.668 $\pm$ 0.812            & 0.184 $\pm$ 0.170            & 0.520 $\pm$ 0.026                                                           & 0.996 $\pm$ 0.048                                                                                   \\
Binary                    & -0.114 $\pm$ 2.147           & -0.191 $\pm$ 3.435           & 0.227 $\pm$ 1.241            & 0.495 $\pm$ 0.077                                                           & 1.002 $\pm$ 0.077                                                                                   \\ \bottomrule
\end{tabular}

\end{table*}

\begin{table*}[htbp]
\centering
\caption{Cross-Validation Performance -- Rescaled to Target Volatility}
\label{table:crossvalidrescaled}
\begin{tabular}{@{}lllll@{}}
\toprule
\textbf{}                 & \textbf{E[Return]}           & \textbf{Vol.}                & \textbf{\begin{tabular}[c]{@{}l@{}}Downside\\     Deviation\end{tabular}} & \textbf{MDD}                 \\ \midrule
{\ul \textbf{Reference}} &                              &                              &                                                                           &                              \\
Long Only                 & 0.131 $\pm$ 0.142            & 0.154 $\pm$ 0.001            & 0.104 $\pm$ 0.014                                                         & 0.304 $\pm$ 0.113            \\
Sgn(Returns)             & 0.186 $\pm$ 0.184            & 0.154 $\pm$ 0.002            & 0.101 $\pm$ 0.012                                                         & 0.194 $\pm$ 0.126            \\
MACD                      & 0.140 $\pm$ 0.166            & 0.154 $\pm$ 0.002            & 0.105 $\pm$ 0.010                                                         & 0.243 $\pm$ 0.129            \\ \midrule
{\ul \textbf{Linear}}     &                              &                              &                                                                           &                              \\
Sharpe                    & 0.182 $\pm$ 0.273            & 0.155 $\pm$ 0.003            & 0.105 $\pm$ 0.007                                                         & 0.232 $\pm$ 0.175            \\
Ave. Returns              & 0.127 $\pm$ 0.141            & 0.154 $\pm$ 0.003            & 0.101 $\pm$ 0.009                                                         & 0.318 $\pm$ 0.177            \\
MSE                       & 0.170 $\pm$ 0.189            & 0.154 $\pm$ 0.003            & \textbf{0.099  $\pm$ 0.006*}                                              & 0.256 $\pm$ 0.221            \\
Binary                    & 0.049 $\pm$ 0.170            & 0.155 $\pm$ 0.002            & 0.104 $\pm$ 0.013                                                         & 0.351 $\pm$ 0.114            \\ \midrule
{\ul \textbf{MLP}}        &                              &                              &                                                                           &                              \\
Sharpe                    & 0.271 $\pm$ 0.375            & 0.154 $\pm$ 0.008            & 0.104 $\pm$ 0.000                                                         & 0.186 $\pm$ 0.259            \\
Ave. Returns              & 0.233 $\pm$ 0.270            & 0.154 $\pm$ 0.003            & 0.101 $\pm$ 0.010                                                         & 0.194 $\pm$ 0.277            \\
MSE                       & 0.148 $\pm$ 0.178            & 0.154 $\pm$ 0.003            & 0.100 $\pm$ 0.009                                                         & 0.268 $\pm$ 0.262            \\
Binary                    & -0.011 $\pm$ 0.117           & 0.154 $\pm$ 0.002            & 0.102 $\pm$ 0.018                                                         & 0.377 $\pm$ 0.221            \\ \midrule
{\ul \textbf{WaveNet}}    &                              &                              &                                                                           &                              \\
Sharpe                    & 0.131 $\pm$ 0.103            & 0.154 $\pm$ 0.002            & 0.104 $\pm$ 0.009                                                         & 0.254 $\pm$ 0.164            \\
Ave. Returns              & 0.142 $\pm$ 0.196            & 0.154 $\pm$ 0.002            & 0.103 $\pm$ 0.003                                                         & 0.262 $\pm$ 0.204            \\
MSE                       & 0.087 $\pm$ 0.150            & \textbf{0.153  $\pm$ 0.003*} & 0.101 $\pm$ 0.009                                                         & 0.307 $\pm$ 0.247            \\
Binary                    & -0.030 $\pm$ 0.099           & 0.155 $\pm$ 0.001            & 0.105 $\pm$ 0.006                                                         & 0.485 $\pm$ 0.283            \\ \midrule
{\ul \textbf{LSTM}}       &                              &                              &                                                                           &                              \\
Sharpe                    & \textbf{0.435  $\pm$ 0.342*} & 0.155 $\pm$ 0.002            & 0.108 $\pm$ 0.012                                                         & \textbf{0.164  $\pm$ 0.077*} \\
Ave. Returns              & 0.157 $\pm$ 0.202            & \textbf{0.153  $\pm$ 0.002*} & 0.102 $\pm$ 0.011                                                         & 0.285 $\pm$ 0.196            \\
MSE                       & 0.087 $\pm$ 0.091            & 0.154 $\pm$ 0.003            & 0.100 $\pm$ 0.006                                                         & 0.310 $\pm$ 0.130            \\
Binary                    & -0.008 $\pm$ 0.332           & 0.155 $\pm$ 0.002            & 0.100 $\pm$ 0.009                                                         & 0.428 $\pm$ 0.495            \\ \bottomrule
\end{tabular}

\begin{tabular}{@{}llllll@{}}
\\\\
\toprule
\textbf{}                 & \textbf{Sharpe}              & \textbf{Sortino}             & \textbf{Calmar}              & \textbf{\begin{tabular}[c]{@{}l@{}}Fraction of \\ $+$ve   Returns\end{tabular}} & \textbf{\begin{tabular}[c]{@{}l@{}}$\mathbf{\frac{\text{Ave.  P}}{\text{Ave. L}}}$\end{tabular}} \\ \midrule
{\ul \textbf{Reference}} &                              &                              &                              &                                                                             &                                                                                                     \\
Long Only                 & 0.847 $\pm$ 0.915            & 1.287 $\pm$ 1.475            & 0.445 $\pm$ 0.579            & 0.546 $\pm$ 0.025                                                           & 0.958 $\pm$ 0.164                                                                                   \\
Sgn(Returns)             & 1.213 $\pm$ 1.205            & 1.856 $\pm$ 1.944            & 1.098 $\pm$ 1.658            & 0.543 $\pm$ 0.061                                                           & 1.028 $\pm$ 0.070                                                                                   \\
MACD                      & 0.911 $\pm$ 1.086            & 1.361 $\pm$ 1.733            & 0.643 $\pm$ 0.958            & 0.532 $\pm$ 0.030                                                           & 1.023 $\pm$ 0.074                                                                                   \\ \midrule
{\ul \textbf{Linear}}     &                              &                              &                              &                                                                             &                                                                                                     \\
Sharpe                    & 1.176 $\pm$ 1.772            & 1.752 $\pm$ 2.615            & 1.060 $\pm$ 2.376            & 0.544 $\pm$ 0.042                                                           & 1.025 $\pm$ 0.139                                                                                   \\
Ave. Returns              & 0.826 $\pm$ 0.914            & 1.287 $\pm$ 1.504            & 0.471 $\pm$ 0.777            & 0.530 $\pm$ 0.009                                                           & 1.016 $\pm$ 0.116                                                                                   \\
MSE                       & 1.101 $\pm$ 1.220            & 1.729 $\pm$ 2.037            & 0.890 $\pm$ 1.787            & 0.540 $\pm$ 0.024                                                           & 1.022 $\pm$ 0.116                                                                                   \\
Binary                    & 0.321 $\pm$ 1.105            & 0.509 $\pm$ 1.720            & 0.169 $\pm$ 0.585            & 0.506 $\pm$ 0.027                                                           & 1.031 $\pm$ 0.127                                                                                   \\ \midrule
{\ul \textbf{MLP}}        &                              &                              &                              &                                                                             &                                                                                                     \\
Sharpe                    & 1.757 $\pm$ 2.405            & 2.623 $\pm$ 3.626            & 2.091 $\pm$ 3.474            & 0.554 $\pm$ 0.063                                                           & 1.085 $\pm$ 0.176                                                                                   \\
Ave. Returns              & 1.516 $\pm$ 1.764            & 2.336 $\pm$ 2.923            & 1.771 $\pm$ 2.889            & 0.553 $\pm$ 0.043                                                           & 1.038 $\pm$ 0.141                                                                                   \\
MSE                       & 0.960 $\pm$ 1.163            & 1.510 $\pm$ 1.927            & 0.864 $\pm$ 1.999            & 0.525 $\pm$ 0.025                                                           & 1.059 $\pm$ 0.103                                                                                   \\
Binary                    & -0.071 $\pm$ 0.756           & -0.140 $\pm$ 1.133           & 0.000 $\pm$ 0.372            & 0.500 $\pm$ 0.042                                                           & 0.991 $\pm$ 0.072                                                                                   \\ \midrule
{\ul \textbf{WaveNet}}    &                              &                              &                              &                                                                             &                                                                                                     \\
Sharpe                    & 0.849 $\pm$ 0.663            & 1.270 $\pm$ 1.060            & 0.575 $\pm$ 0.680            & 0.535 $\pm$ 0.022                                                           & 1.000 $\pm$ 0.121                                                                                   \\
Ave. Returns              & 0.920 $\pm$ 1.271            & 1.376 $\pm$ 1.915            & 0.738 $\pm$ 1.591            & 0.543 $\pm$ 0.059                                                           & 0.979 $\pm$ 0.066                                                                                   \\
MSE                       & 0.565 $\pm$ 0.972            & 0.854 $\pm$ 1.513            & 0.364 $\pm$ 0.665            & 0.527 $\pm$ 0.033                                                           & 0.986 $\pm$ 0.081                                                                                   \\
Binary                    & -0.196 $\pm$ 0.641           & -0.298 $\pm$ 0.946           & -0.044 $\pm$ 0.206           & 0.499 $\pm$ 0.011                                                           & 0.974 $\pm$ 0.067                                                                                   \\ \midrule
{\ul \textbf{LSTM}}       &                              &                              &                              &                                                                             &                                                                                                     \\
Sharpe                    & \textbf{2.803  $\pm$ 2.195*} & \textbf{4.084  $\pm$ 3.469*} & \textbf{2.887  $\pm$ 3.030*} & \textbf{0.593  $\pm$ 0.054*}                                                & \textbf{1.106  $\pm$ 0.216*}                                                                        \\
Ave. Returns              & 1.023 $\pm$ 1.312            & 1.564 $\pm$ 2.131            & 0.706 $\pm$ 1.440            & 0.547 $\pm$ 0.039                                                           & 0.980 $\pm$ 0.127                                                                                   \\
MSE                       & 0.563 $\pm$ 0.580            & 0.865 $\pm$ 0.901            & 0.284 $\pm$ 0.269            & 0.520 $\pm$ 0.026                                                           & 1.014 $\pm$ 0.016                                                                                   \\
Binary                    & -0.050 $\pm$ 2.152           & -0.122 $\pm$ 3.381           & 0.190 $\pm$ 1.152            & 0.495 $\pm$ 0.077                                                           & 1.012 $\pm$ 0.048                                                                                   \\ \bottomrule
\end{tabular}
\end{table*}

\newpage
\pagenumbering{gobble}

\end{document}